\documentclass[twoside]{article}

\usepackage[accepted]{aistats2024}
\usepackage[utf8]{inputenc} 
\usepackage[T1]{fontenc}    
\usepackage{booktabs}       
\usepackage{amsfonts,amsmath,bm}       
\usepackage{nicefrac}       
\usepackage{microtype}      
\usepackage{xcolor}         
\usepackage{multirow}
\usepackage{multicol}
\usepackage{comment}

\usepackage[style=numeric,maxcitenames=1]{biblatex} 
\usepackage{mathtools}
\usepackage{graphicx}
\usepackage{enumitem} 
\usepackage{wrapfig}
\usepackage{placeins} 
\usepackage{caption}
\usepackage{subcaption} 
\usepackage{csquotes}
\usepackage{pgfplots}
\pgfplotsset{compat=newest}
\usepgfplotslibrary{groupplots}
\usepgfplotslibrary{polar}
\usepgfplotslibrary{smithchart}
\usepgfplotslibrary{statistics}
\usepgfplotslibrary{dateplot}
\usepgfplotslibrary{ternary}
\usetikzlibrary{arrows.meta}
\usetikzlibrary{backgrounds}
\usepgfplotslibrary{patchplots}
\usepgfplotslibrary{fillbetween}
\pgfplotsset{%
	layers/standard/.define layer set={%
		background,axis background,axis grid,axis ticks,axis lines,axis tick labels,pre main,main,axis descriptions,axis foreground%
	}{
		grid style={/pgfplots/on layer=axis grid},%
		tick style={/pgfplots/on layer=axis ticks},%
		axis line style={/pgfplots/on layer=axis lines},%
		label style={/pgfplots/on layer=axis descriptions},%
		legend style={/pgfplots/on layer=axis descriptions},%
		title style={/pgfplots/on layer=axis descriptions},%
		colorbar style={/pgfplots/on layer=axis descriptions},%
		ticklabel style={/pgfplots/on layer=axis tick labels},%
		axis background@ style={/pgfplots/on layer=axis background},%
		3d box foreground style={/pgfplots/on layer=axis foreground},%
	},
}

\usepackage{url}            
\usepackage{hyperref}       

\newcommand{\Rm}{\mathcal{R}^{-}}
\newcommand{\Ip}{\mathcal{I}^{+}}
\renewcommand{\Im}{\mathcal{I}^{-}}
\DeclareMathOperator*{\indep}{\perp \!\!\! \perp}
\renewcommand{\H}{\operatorname{H}}
\newcommand{\II}{\mathbb{I}}
\DeclareMathOperator{\tr}{tr}
\DeclareMathOperator{\diag}{diag}
\DeclareMathOperator{\MI}{MI}
\newcommand{\trans}{^{\top}}
\newcommand{\mtrans}{^{-\top}}
\renewcommand{\Re}{\mathbb{R}}
\newcommand{\I}{\mathcal{I}}
\newcommand{\R}{\mathcal{R}}

\newcommand{\F}{\mathcal{F}}
\newcommand{\nystrom}{Nystr\"{o}m }

\DeclareMathOperator*{\argmax}{arg\,max}

\newcommand{\one}{1}
\newcommand{\zero}{0}
\newcommand{\Cov}{\operatorname{Cov}}
\newcommand{\Var}{\operatorname{Var}}
\DeclarePairedDelimiterX{\norm}[1]{\lVert}{\rVert}{#1}
\newcommand{\dsname}[1]{\text{#1}}
\newcommand{\est}[1]{\hat{#1}}
\newcommand{\perm}[1]{\tilde{#1}}

\newcommand*\phantomrel[1]{\mathrel{\phantom{#1}}}

\graphicspath{{fig/}{fig/selection_visualization}}

\makeatletter
\newcommand\thefontsize{The current font size is: \f@size pt and \f@linewidth}
\makeatother

\title{A Pivotal Moment for Gaussian Processes}

\definecolor{snsblue}{rgb}{0.121569,0.466667,0.705882}
\definecolor{snsgreen}{rgb}{0.172549,0.627451,0.172549}
\definecolor{snsorange}{rgb}{1.000000,0.498039,0.054902}
\definecolor{snsred}{rgb}{0.839216,0.152941,0.156863}
\definecolor{snspurple}{rgb}{0.580392,0.403922,0.741176}

\begin{document}

%

%

\twocolumn[%
\aistatstitle{Novel Pivoted Cholesky Decompositions\\for Efficient Gaussian Process Inference}%
\aistatsauthor{Filip de Roos$^*$ \And Fabio Muratore$^*$}%
\aistatsaddress{Robert Bosch GmbH (Bosch Center for Artificial Intelligence)}%
]



\begin{abstract}
    The Cholesky decomposition is a fundamental tool for solving linear systems with symmetric and positive definite matrices which are ubiquitous in linear algebra, optimization, and machine learning.
    Its numerical stability can be improved by introducing a pivoting strategy that iteratively permutes the rows and columns of the matrix.
    The order of pivoting indices determines how accurately the intermediate decomposition can reconstruct the original matrix, thus is decisive for the algorithm's efficiency in the case of early termination.
    Standard implementations select the next pivot from the largest value on the diagonal.
    In the case of Bayesian nonparametric inference, this strategy corresponds to greedy entropy maximization, which is often used in active learning and design of experiments.
    We explore this connection in detail and deduce novel pivoting strategies for the Cholesky decomposition.
    The resulting algorithms are more efficient at reducing the uncertainty over a data set, can be updated to include information about observations, and additionally benefit from a tailored implementation.
    We benchmark the effectiveness of the new selection strategies on two tasks important to Gaussian processes: sparse regression and inference based on preconditioned iterative solvers.
    Our results show that the proposed selection strategies are either on par or, in most cases, outperform traditional baselines while requiring a negligible amount of additional computation.
\end{abstract}

\section{Introduction}
Gaussian processes (GPs) are a popular nonparametric framework for probabilistic modeling with widespread application in scenarios where the uncertainty of predictions is important.
For GPs, learning is achieved by conditioning a latent function on observations which are often considered to be corrupted by a normally distributed error term.
Conditioning a GP on observations requires the solution of a linear system of $N$ equations, where $N$ is the number of data points and the matrix is symmetric and positive definite (SPD).
A standard approach to solve this system of equations is to perform the Cholesky decomposition~\cite{golub2013matrix} on the $N\!\times\!N$ matrix and use the resulting factorization as a direct solution method.
However, the Cholesky decomposition overall runtime scales cubically with the matrix dimensionality $\mathcal{O}(N^3)$, which for GPs corresponds to the size of the data set.
Hence, the applicability of GPs is sensitive to the size of the data set which can be prohibitive for large data sets such that approximation techniques must be considered.
Pivoting is a state-of-the-art method to increase the numerical stability of Cholesky decompositions~\cite{golub2013matrix}.
In short, this technique iteratively permutes the rows and columns of the matrix at every step of the decomposition according to some criterion.
This criterion, i.e., the pivoting strategy, becomes particularly important when the Cholesky decomposition is stopped early to yield a low-rank approximation.

In this work, we study the intermediate quantities that are used within the pivoted Cholesky decomposition and find a remarkable connection to inducing point methods as well as active learning strategies when applied to GPs. 
We then augment the pivoted Cholesky decomposition with novel pivoting strategies.
By altering the pivoting strategy, we can generalize the procedure to better capture statistically more relevant quantities which results in a more efficient data selection for downstream applications.
The ubiquity of the Cholesky decomposition, especially in statistics, e.g. sampling sigma points and Gaussian inference means that our findings easily translate to other fields and applications beyond GPs.
Moreover, due to its cubic scaling, the decomposition often constitutes a computational bottleneck of numerical algorithms.
Therefore, efficient approximations are greatly coveted and even seemingly minor improvements can have significant benefits~\cite{fawzi2022alphatensor}. 

\textbf{Contributions:}
\begin{enumerate}[leftmargin=*, nosep]
    \item We identify the standard pivoted Cholesky decomposition as a greedy active learning strategy based on entropy maximization that can be terminated prematurely to obtain a low-rank matrix approximation of the matrix (Section~\ref{sec:entropy_selection}).
    \item Based on this analysis, we motivate two new pivoting strategies for the Cholesky decomposition: the projected covariance (PCov, Section~\ref{sec:pcov}) which incorporates information about the full covariance matrix, and the weighted projected covariance (WPCov, Section~\ref{sec:wpcov}) which additionally incorporates information about the observations.
    Both algorithms only require a negligible computational overhead. 
    \item In our experiments, we demonstrate how the introduced pivoting strategies can be used as an efficient and generic preconditioner to improve the convergence rate of iterative linear solvers 
    (Section~\ref{sec:cg_convergence}).
    Furthermore, we show that both methods are beneficial for sparse GP regression tasks as they require fewer inducing points to obtain the same approximation quality (Section~\ref{sec:gp_regression}).
    \item Finally, we implement all pivoting strategies in Julia as an alternative to the standard pivoted Cholesky routine, along with an efficient data structure for preconditioners which decreases the computation time for the preconditioning step by up to one order of magnitude (Appendix~\ref{asec:preconditioner}).
\end{enumerate}

\section{Background and Notation}
\label{sec:background}
For simplicity, we restrict the analysis to a regression task where a data set consists of $N$ input-output pairs
$\{x_i,y_i\}^{N}_{i=1}$ stored as the matrix $X\in\Re^{N \times D}$ and vector $y\in \Re^N$, where $D$ is the dimension of the input. %
The data will be further partitioned into a set of included points $\I \subset \{1, 2, \dots, N\}$ of size $M$, and the remaining set $\R = \{1, 2, \dots, N\} \setminus \I$ of size $N-M$.
We also define the permuted data set where data has been reordered as $\perm{X} = [X_{\I}, X_{\R}]$ and $\perm{y} = [y_{\I}, y_{\R}]$.
For brevity, we abuse the notation and let the partitions both refer to indices and data points because $K_{\I \I} = K_{X_{\I} X_{\I}}$.

\subsection{Gaussian Processes}
A GP is a distribution over functions where realizations share a multivariate normal distribution. %
The most important component of a GP is the bivariate kernel function $k(\cdot,\cdot)$ which encodes the covariance between observations as well as properties of the latent function.
In this paper, we restrict ourselves to stationary kernels with non-negative covariance, and in particular explore the exponentiated quadratic (EQ) kernel with automatic relevance determination (ARD)~\cite[Ch. 5.1][]{rasmussen2006gaussian} $k(x,x') = \theta \exp(-(x-x')\trans\Lambda (x-x')/2)$.
If a prior over a latent function $p(f)\sim \mathcal{GP}(0, k)$ is evaluated at points $X$, we obtain $p(f_X)= \mathcal{N}(0, K_{XX})$, where $K_{XX}\in \Re^{N \times N}$ is a symmetric and semi-definite matrix containing all the pair-wise evaluations of $k(\cdot, \cdot)$ applied to the points in $X$.
If observations $y$ of $f$ are corrupted by independent and homogeneous Gaussian noise $p(y_i \mid f(x_i)) = \mathcal{N}(f(x_i),\sigma^2)$, then by the conjugacy of Gaussian distributions we get $p(y) = \mathcal{N}(f_X, G_{XX})$, where $G_{XX} = K_{XX} + \sigma^2 I$ is the Gram matrix, which is SPD.
Conditioning a GP on noisy observations in this setting requires the solution of the linear system ${G_{XX} \alpha = y}$, where $\alpha$ is an intermediate quantity to evaluate the posterior mean at a new point $f_{*} = K_{*X} \alpha$.
The \enquote{standard} inference procedure makes use of the Cholesky decomposition~\cite[][Ch.~2]{rasmussen2006gaussian}.
However, the overall $\mathcal{O}(N^3)$ computational and $\mathcal{O}(N^2)$ storage costs are prohibitive for many real-world applications so approximations are required.
To alleviate this issue, plenty of sparse approximations have been proposed~\cite[][Ch.~8]{rasmussen2006gaussian}~and~\cite{quinonero2005unifying,schreiter2016efficient}. 

\subsection{Sparse Gaussian Processes}
\label{sec:sparse_gps}
Significant progress has been made towards improving the training efficiency of kernel methods by approximating either the kernel function~\cite{rahimi2007rff}, the kernel matrix~\cite{williams2000nystrom}, the conditional dependency~\cite{quinonero2005unifying}, or the posterior distribution~\cite{titsias2009variational}.
In this work, we focus on the variational free energy (VFE) approximation~\cite{bauer2016variational,titsias2009variational} for inducing points selected from the training set~\cite{burt19convergence}.

The VFE training objective makes use of the \nystrom approximation~\cite{williams2000nystrom} which is a low-rank rank approximation of the full covariance matrix
\begin{equation}
    \est{K}_{XX} = K_{X \I}K_{\I \I}^{-1}K_{\I X} \approx K_{XX},
    \label{eq:nystrom}
\end{equation}
with $K_{X \I}\in \Re^{N \times M} = \Cov[f_{X}, f_{\I}]$, and $K_{\I \I} \in \Re^{M \times M} = \Cov[f_{I}, f_{\I}]$ being the covariance matrices of the latent function evaluated at the respective data points.
The original implementation~\cite{williams2000nystrom} samples the points $\I$ uniformly from the set $\{1,\dots,N\}$, yielding an unbiased estimator. 
Depending on the task, it is better to employ a (biased) sampling strategy specifically targeting a statistical quantity of interest~\cite{burt19convergence,chen2018piv_dpp}.

Optimizing the VFE objective is equivalent to minimizing a penalized version of the negative log marginal likelihood (NLML) which consists of a constant and three competing terms that each encode a different behavior~\cite{bauer2016variational}
\begin{align}
    -\F &= \frac{N}{2} \log \left(2\pi \right)  &&\text{(constant)} \notag\\ 
    &+ \frac{1}{2} y \trans \left(\est{K}_{XX} + \sigma^2 I \right)^{-1} y &&\text{(data fit)} \notag\\
    &+ \frac{N}{2} \log \left| \est{K}_{XX} + \sigma^2 I \right| &&\text{(complexity)} \notag\\
    &+ \frac{1}{\sigma^2} \tr \left(K_{XX} - \est{K}_{XX} \right). &&\text{(penalty)}
    \label{eq:elbo}
\end{align}
The trace penalty term ensures that $\F$ in~\eqref{eq:elbo} constitutes a lower bound to the evidence (ELBO) and $\est{K}_{XX}$ is the \nystrom approximation~\eqref{eq:nystrom}.
The value of the NLML strongly depends on the selection of inducing points $X_{\I}$ which can be random~\cite{williams2000nystrom}, continuously optimized along with kernel hyperparameters~\cite{chang2023piv_ind,hensman2013bigdata,titsias2009variational}, or sampled according to a greedy selection criterion\cite{burt19convergence,lawrence2002ivm,quinonero2005unifying,schreiter2016efficient,seeger2003fast,smola2000greedymatrix}.
The greedy selection strategies often have a computational cost of $\mathcal{O}(M^2N)$ where $M$ is the number of inducing points, thus are employed to counteract the combinatorial explosion of possible partitions\cite{ko1995np_hard}.

\subsection{Cholesky Decomposition}
\label{sec:cholesky}
The Cholesky decomposition~\cite[][Ch.~4]{golub2013matrix} is frequently used as a direct solution method for solving a system of linear equations when the matrix in question is SPD\footnote{A matrix $A\in \Re^{N \times N}$ is said to be SPD if $A=A\trans$ and $v\trans A v>0\, \forall\, v\in \Re^{N}\setminus \{0\}$.}.
The result is a lower-triangular matrix\footnote{Some implementations instead return an upper-triangular matrix $U$ which is the transpose of $L$.} $L$ such that ${ K = L L\trans }$.
Performing the full decomposition has a computational cost of $\mathcal{O}(N^3)$, where $N$ is the dimension of the matrix.
Once the decomposition is available, the solution is computed in $\mathcal{O}(N^2)$ by using the triangular structure and performing a forward and backward substitution with $L$~\cite[][Ch.~3.1]{golub2013matrix}.

A partial Cholesky decomposition refers to the result of a prematurely terminated procedure.
After $M$ iterations, we have $\I=\{1,\dots,M\}$, $\R=\{M+1,\dots,N\}$, and the intermediate matrix $L_{X \I} \in \Re^{N \times M}$ partitioned as $L_{X \I} = [L_{\I \I}, L_{\R \I}]\trans$.
The top block $L_{\I \I}$ is the lower-triangular Cholesky decomposition of $K_{\I \I}$, which for kernel methods corresponds to the Gram matrix of the first $M$ data points.
The lower block of $L_{X \I}$ is $L_{\R \I} = (L_{\I \I}^{-1 } K_{\I \R})\trans$.
Further inspection of $L_{X \I}$ reveals a close connection to the \nystrom approximation \eqref{eq:nystrom}
\begin{align}
  L_{X \I}L_{X \I}\trans 
  &= L_{X \I}L_{\I \I}\trans (L_{\I \I}L_{\I \I}\trans)^{-1} L_{\I \I}L_{X \I}\trans \label{eq:L_outer} \notag\\
  &= K_{X \I} (K_{\I \I})^{-1} K_{X \I}\trans = \est{K}_{X X}.
\end{align}

\subsection{Pivoted Cholesky Decomposition}
The key difference between the normal and the pivoted variant is the latter's possibility to permute the rows and columns of the matrix.
This makes the algorithm numerically more stable because division by numbers close to zero can be avoided at early stages in the procedure~\cite[][Sec.~4.2.9]{golub2013matrix}.
At iteration $m$ a pivot index $\pi_m \in m, \dots, N$ is chosen, and the corresponding rows and columns are swapped ($m\leftrightarrow \pi_m$). The resulting decomposition is then 
\begin{equation}
    L_{\perm{X} \perm{X}}L_{\perm{X} \perm{X}}\trans = K_{\perm{X} \perm{X}} = P\trans K_{X X} P,
\end{equation} 
where $P$ is a sparse permutation matrix with $P_{i\pi_i}=1$ for $i=1,\dots,N$ and $\perm{X} = [X_{\I}, X_{\R}]$ is the permuted data set.
For efficiency reasons, the matrix $P$ is never built explicitly, instead, the vector of pivot indices $\pi$ is used.
The pivoted Cholesky decomposition has virtually the same run time as the standard version but requires the additional storage of 1 or 2 vectors of length $N$ depending on the implementation.

\subsection{Active Learning}
\label{sec:active_learning}
Active learning is a form of sequential decision-making with the goal of learning a latent function, e.g. a dynamics model.
By employing GPs, active learning methods can incorporate prior knowledge about the function as well as encode known covariance structure to help guide the exploration.
Active learning is used for classification as well for regression tasks~\cite{Joshi09,Zhang16,Gramacy20}, some of which additionally consider safety~\cite{schreiter2015safeALDE,Zimmer18}.
\citeauthor{settles2009alsurvey}~\cite{settles2009alsurvey} provides an overview of the topic, along with experiments highlighting the benefits and potential drawbacks of employing active learning in different scenarios.
Moreover, active learning is intimately connected to design of experiments where the next experiment to conduct is chosen to be maximally informative based on all of the previous experiments.
The notion of informativeness can take on a different meaning depending on the final goal, thus leading to different strategies~\cite[][Ch. 2.1]{fedorov-hackl} of which D-optimal (optimization w.r.t. the determinant) and A-optimal (optimization w.r.t. the trace) will be of particular interest for the remainder of this paper.

\section{Theory}
\label{sec:theory}
In the following, we set out how the standard pivoted Cholesky decomposition can be seen as a greedy form of D-optimal design (Section~\ref{sec:entropy_selection}).
After discussing its shortcomings, we explore how ideas from sparse GP inference and active learning come together to create novel pivoting strategies for the Cholesky decomposition (Sections~\ref{sec:pcov}~and~\ref{sec:wpcov}).

\subsection{Entropy-Maximizing Selection Strategy}
\label{sec:entropy_selection}
State-of-the-art linear algebra libraries\footnote{\hyperlink{https://netlib.org/lapack/explore-html/da/dba/group__double_o_t_h_e_rcomputational_ga31cdc13a7f4ad687f4aefebff870e1cc.html}{dpstrf} in LAPACK and \hyperlink{https://www.netlib.org/linpack/dchdc.f}{dchdc} in LINPACK.} choose the next pivoting element based on the maximum diagonal value of the Schur complement to $K_{\I \I}$ 
\begin{align} 
  \pi_m 
  &= \argmax_{j \in \R} \diag(K_{\perm{X} \perm{X} } - L_{\perm{X} \I}L_{\perm{X} \I}\trans)_{j} \notag\\
  &= \argmax_{j \in \R} \diag(K_{\perm{X} \perm{X} } - K_{\perm{X} \I}K_{\I \I}^{-1}K_{\I \perm{X} })_{j} \notag\\
  &= \argmax_{j \in \R} \diag(K_{\R\R} - K_{\R\I}K_{\I\I}^{-1}K_{\I\R})_{j} \label{eq:variance}
\end{align}
which offers an interesting interpretation in the case of GPs.
From the last expression, we see that the chosen point is the point in $\R$ with the highest variance when conditioned on the observations in $\I$.
Another way to phrase it is the point at which we are the most uncertain or the point with the highest entropy.
The selection in~\eqref{eq:variance} is often referred to as a form of entropy search~\cite[][Ch.~4.3]{mackay1992thesis}.
Since the entropy of a Gaussian random variable is proportional to the log determinant of the covariance matrix~\cite[][Appendix~B]{bishop2006prml}, it should come as no surprise that it is also related to a greedy form of D-optimal design~\cite{fedorov-hackl}.
We derive this relationship in Appendix~\ref{asec:d-optimal}.

The variance selection \eqref{eq:variance} has found widespread application within machine learning and statistics, often under a different name. 
In the case of stationary kernels where the covariance depends on the relative proximity of data points, the selection strategy in \eqref{eq:variance} has been employed under the names of uncertainty sampling, space-filling, and relative entropy minimization~\cite{ko1995np_hard,krause2007differentialentropy,schreiter2015safeALDE}.
Moreover, it also is closely linked to determinantal point processes either as an approximation or initialization for further sampling~\cite{chen2018piv_dpp,hennig2016dpp}.
\citeauthor{lawrence2002ivm}~\cite{lawrence2002ivm} built a subset of data for use in kernel ridge regression with the variance selection and inadvertently implemented the pivoted Cholesky algorithm in the process~\cite[][Sec.~8.2.1]{rasmussen2006gaussian}. 
Several of the aforementioned frameworks can thus be realized with a standard pivoted Cholesky decomposition when a stationary kernel is used and does not, per se, necessitate the development of a new algorithm.

A potential downside of this selection rule is that for stationary kernels the point with the maximum variance is the one that is the furthest away from the observed points $X_{\I}$~\cite[][Ch.~4.3]{mackay1992thesis}.
In principle, this means that the selection would in the beginning favor points on the edges of the data or outliers.
This effect is aggravated in higher dimensions where the number of outlier directions in a hypercube scales exponentially with the dimension $\mathcal{O}(2^D)$. 
In practice, however, the expected behavior is not observed for kernels with a rapidly decaying kernel function such as the exponentiated quadratic.
Considering this type of kernel, a variance reduction more than a few characteristic lengthscales away is below the available numerical precision, thus the $\argmax$ operator in~\eqref{eq:variance} selects a point that lies in the interior of the data rather than an outlier, which would be the true maximizer.
This behavior is displayed for the Var-selection in Figure~\ref{fig:selection_strategies_overview} as well as in its extended version, Figure~\ref{afig:selection_strategies_all} within the Appendix~\ref{asec:selection_visualization}.

The Var-selection~\eqref{eq:variance} is mainly focused on minimizing the complexity term in~\eqref{eq:elbo}, whereas the remaining terms often have a larger influence on the NLML.
This can be seen from the subpar performance of the Var baseline across all data sets for the experiments we describe in Section~\ref{sec:gp_regression}.
To counteract the undesired behavior, we provide two new pivoting strategies that instead focus on the trace penalty (Section~\ref{sec:pcov}) or additional the data fit (Section~\ref{sec:wpcov}) term.


\subsection{PCov - Approximate Trace Minimization}
\label{sec:pcov}
In active learning, we often want to learn as much information as possible about a function.
In such a setting it becomes useful to select points $X_{\I}$ to maximally reduce the uncertainty of the latent function at the remaining points $X_{\R}$.
The goal is then to minimize $\tr( K_{XX} - \est{K}_{XX} ) = \tr( K_{\perm{X}\perm{X}} - \est{K}_{\perm{X}\perm{X}} )$ which also appears as the penalty term in~\eqref{eq:elbo}.

Selecting the point in $\R$ that maximally reduces the uncertainty of the other points is known as a greedy form of A-optimal design (Appendix~\ref{asec:a-optimal}). 
For one-step look-ahead, there exists a closed-form expression, but it requires one MVP per iteration which can be prohibitive for large covariance matrices.
To alleviate this computational cost, we construct a new selection strategy called projected covariance (PCov) which selects a new point at iteration $m$ according to
\begin{align}
    \pi_m 
	&= \argmax\limits_{j\geq m} \, \norm[\big]{ \left(K_{\perm{X} \perm{X}} - K_{\perm{X}\I}(K_{\I\I})^{-1}K_{\I \perm{X}} \right) \one_{N} }^2_j \notag\\
	&= \argmax\limits_{j\geq m} \, \norm[\big]{ s^{*}_{\perm{X}} - s^{(m)}_{\perm{X}} }^2_j.  \label{eq:pcov}
\end{align}
Here, $s^{*}_{\perm{X}}= K_{\perm{X}\perm{X}}\one_{N}$ is a fixed vector calculated beforehand with one MVP.
The second vector is initialized as $s^{(1)}_{\perm{X}}=0_{N}$ and then iteratively updated to contains
\begin{equation}
  s^{(m+1)}_{\perm{X}} = K_{\perm{X}\I}(K_{\I\I})^{-1}K_{\I \perm{X}} \one_{N},
  \label{eq:pcov_update}
\end{equation}
where $\I$ consists of all $m$ indices that have been collected so far.
By caching previous computations and using intermediate results from the Cholesky routine we can turn the $\mathcal{O}(MN)$ update operation of \eqref{eq:pcov_update} into an $\mathcal{O}(N)$ operation which is on par with the standard pivoted Cholesky routine (Appendix~\ref{asec:pcov_implementation_details_comparison}).
In Appendix~\ref{asec:trace_estimation}, we show that this selection constitutes an upper bound on the possible trace reduction and also influences a lower bound of the trace reduction.
It is also possible to relate this selection to a simplistic approximation of the mutual information (MI)
that only considers the pairwise information between random variables and not the whole ensemble.

\subsection{WPCov - Including Observations}
\label{sec:wpcov}
The PCov selection strategy is motivated by a variance reduction of the latent function at the data points in $X$.
However, for regression tasks, it is more desirable to minimize the $\ell_2$-norm $\norm{ \perm{y} - f_{\perm{X} \mid \I} }^2 = \norm{ \perm{y} - K_{\tilde{X}\I}\alpha_{\I} }^2$, which appears as the data fitting term in~\eqref{eq:elbo}.
A cheap and straightforward approach to this end is to select new points based on the maximum squared error ${ \pi_m = \argmax \norm{ \perm{y} - f_{\perm{X} \mid \I} }_j^2 }$, which was introduced as maximum error (ME) search by~\citeauthor{schreiter2016efficient}~\cite{schreiter2016efficient}.
A decisive downside of this selection is that the point with the maximum $\ell_2$ error is not necessarily the point that would lead to the overall best decrease in error.

A more suitable option for $\ell_2$ optimization, known as matching pursuit or forward regression algorithms~\cite{ament2021sparse}, iteratively updates the residual and selects the next point based on the magnitude of the gradient of the $\ell_2$-norm $\nabla_{\alpha} \norm{ \perm{y} - K_{\tilde{X}\I}\alpha_{\I} }^2 = -K_{\perm{X} \perm{X}} ( \perm{y} - K_{\tilde{X}\I}\alpha_{\I} )$.
Expressed similarly to \eqref{eq:pcov} a matching pursuit selection thus reads $\pi_m = \argmax |K_{\perm{X} \perm{X}} (\perm{y} - K_{\tilde{X}\I}\alpha_{\I} )|_j$.  
Such algorithms can efficiently decrease the $\ell_2$-norm of the residual, but similar to greedy A-optimal design, they also require one MVP per iteration which can be prohibitive for large matrices, i.e., large data sets.

We instead propose to adapt the PCov algorithm in order to reduce the residual's $\ell_2$-norm by weighting each point differently, creating the weighted projected covariance (WPCov) selection strategy.
The adaptation manifests itself by defining the vectors containing the summary statistics to use the initial residual $(\perm{y} - \mu_{\perm{X}})$ instead of the $\one_N$-vector.
The new vectors
\begin{equation}
\begin{split}
    s^{*}_{\perm{X}} &= K_{\perm{X} \perm{X}} (\perm{y} - \mu_{\perm{X}}), \\
    s^{(m+1)}_{\perm{X}} &= K_{\perm{X} \I}(K_{\I \I})^{-1}K_{\I \perm{X}}(\perm{y} - \mu_{\perm{X}}),
\end{split}
\label{eq:wpcov}
\end{equation}
can be used with the same selection criterion~\eqref{eq:pcov} and update rule~\eqref{eq:pcov_update} as PCov.
Hence, the overall computational cost is identical to PCov, i.e., a single MVP upfront plus the usual $\mathcal{O}(NM^2)$.
The behavior of WPCov is similar to a matching pursuit algorithm but with a different interpretation.
WPCov updates the features that are matched with the initial residual such that at iteration $m$ the selection uses the effective features ${ K^{(m)}_{\perm{X} \perm{X}} (\perm{y} - \mu_{\perm{X}}) = (K_{\perm{X} \perm{X}} - K_{\perm{X} \I}(K_{\I \I})^{-1}K_{\I \perm{X}})( \perm{y} - \mu_{\perm{X}}) }$.
In contrast, matching pursuit keeps the features fixed but updates the residual instead.

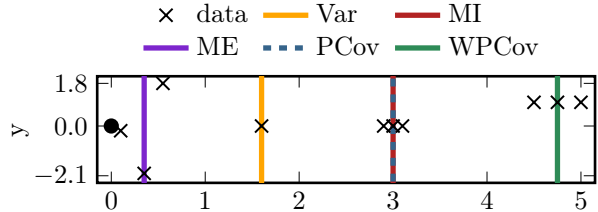
\begin{figure}[t]
\raggedright

\begin{tikzpicture}[/tikz/background rectangle/.style={fill={rgb,1:red,1.0;green,1.0;blue,1.0}, fill opacity={1.0}, draw opacity={1.0}}, show background rectangle]
\begin{axis}[point meta max={nan}, point meta min={nan}, legend cell align={left}, legend columns={3}, title={}, title style={at={{(0.5,1)}}, anchor={south}, font={{\fontsize{10 pt}{13.0 pt}\selectfont}}, color={rgb,1:red,0.0;green,0.0;blue,0.0}, draw opacity={1.0}, rotate={0.0}, align={center}}, legend style={color={rgb,1:red,0.0;green,0.0;blue,0.0}, draw opacity={0.0}, line width={1}, solid, fill={rgb,1:red,1.0;green,1.0;blue,1.0}, fill opacity={1.0}, text opacity={1.0}, font={{\fontsize{10 pt}{13.0 pt}\selectfont}}, text={rgb,1:red,0.0;green,0.0;blue,0.0},
cells={anchor={west}},
at={(0.5, 1.05)}, anchor={south}},
axis background/.style={fill={rgb,1:red,1.0;green,1.0;blue,1.0}, opacity={1.0}}, anchor={north west},
xshift={0.0mm}, yshift={0.0mm}, width={82mm}, height={30mm},
scaled x ticks={false}, xlabel={}, x tick style={color={rgb,1:red,0.0;green,0.0;blue,0.0}, opacity={1.0}}, x tick label style={color={rgb,1:red,0.0;green,0.0;blue,0.0}, opacity={1.0}, rotate={0}}, xlabel style={at={(ticklabel cs:0.5)}, anchor=near ticklabel, at={{(ticklabel cs:0.5)}}, anchor={near ticklabel}, font={{\fontsize{10 pt}{13.0 pt}\selectfont}}, color={rgb,1:red,0.0;green,0.0;blue,0.0}, draw opacity={1.0}, rotate={0.0}}, xmajorgrids={false}, xmin={-0.15}, xmax={5.15},
xticklabels={{0,1,2,3,4,5}},
xtick={{0.0,1.0,2.0,3.0,4.0,5.0}},
xtick align={inside}, xticklabel style={font={{\fontsize{10 pt}{13.0 pt}\selectfont}}, color={rgb,1:red,0.0;green,0.0;blue,0.0}, draw opacity={1.0}, rotate={0.0}}, x grid style={color={rgb,1:red,0.0;green,0.0;blue,0.0}, draw opacity={0.1}, line width={0.5}, solid}, xticklabel pos={left}, x axis line style={color={rgb,1:red,0.0;green,0.0;blue,0.0}, draw opacity={1.0}, line width={1}, solid}, scaled y ticks={false}, ylabel={y}, y tick style={color={rgb,1:red,0.0;green,0.0;blue,0.0}, opacity={1.0}}, y tick label style={color={rgb,1:red,0.0;green,0.0;blue,0.0}, opacity={1.0}, rotate={0}}, ylabel style={at={(ticklabel cs:0.5)}, anchor=near ticklabel, at={{(ticklabel cs:0.5)}}, anchor={near ticklabel}, font={{\fontsize{10 pt}{13.0 pt}\selectfont}}, color={rgb,1:red,0.0;green,0.0;blue,0.0}, draw opacity={1.0}, rotate={0.0}}, ymajorgrids={false},
y label style={yshift=-.5em},
ymin={-2.4}, ymax={2.1},
yticklabels={{$-2.1$,$0.0$,$1.8$}},
ytick={{-2.1,0.0,1.8}},
ytick align={inside}, yticklabel style={font={{\fontsize{10 pt}{13.0 pt}\selectfont}}, color={rgb,1:red,0.0;green,0.0;blue,0.0}, draw opacity={1.0}, rotate={0.0}}, y grid style={color={rgb,1:red,0.0;green,0.0;blue,0.0}, draw opacity={0.1}, line width={0.5}, solid}, yticklabel pos={left}, y axis line style={color={rgb,1:red,0.0;green,0.0;blue,0.0}, draw opacity={1.0}, line width={1}, solid}, colorbar={false}]
    \addplot[color={rgb,1:red,0.0;green,0.0;blue,0.0}, name path={d91fbaa8-26a4-4dbe-8450-0f81039661ab}, only marks, draw opacity={1.0}, line width={0}, solid, mark={x},
    mark size={3.5 pt},
    mark repeat={1}, mark options={color={rgb,1:red,0.0;green,0.0;blue,0.0}, draw opacity={1.0}, fill={rgb,1:red,0.0;green,0.0;blue,0.0}, fill opacity={1.0}, line width={0.75}, rotate={0}, solid}]
        table[row sep={\\}]
        {
            \\
            0.1  -0.2  \\
            0.35  -2.0  \\
            0.55  1.8  \\
            1.6  0.0  \\
            2.9  0.0  \\
            3.0  0.0  \\
            3.1  0.0  \\
            4.5  1.0  \\
            4.75  1.0  \\
            5.0  1.0  \\
        }
        ;
    \addplot[color={rgb,1:red,1.0;green,0.6471;blue,0.0}, name path={e06a284c-0002-4a85-8f93-96e868686194}, draw opacity={1.0}, line width={2}, solid]
        table[row sep={\\}]
        {
            \\
            1.6  -6.36  \\
            1.6  6.36  \\
        }
        ;
    \addplot[color={rgb,1:red,0.698;green,0.1333;blue,0.1333}, name path={98d0c08d-a822-4f74-97c2-4f5f496c9bdb}, draw opacity={1.0}, line width={2}, solid]
        table[row sep={\\}]
        {
            \\
            3.0  -6.36  \\
            3.0  6.36  \\
        }
        ;
    \addplot[color={rgb,1:red,0.4902;green,0.149;blue,0.8039}, name path={207bcf28-9ff3-46a7-be02-fce1ee3e48d2}, draw opacity={1.0}, line width={2}, solid]
        table[row sep={\\}]
        {
            \\
            0.35  -6.36  \\
            0.35  6.36  \\
        }
        ;
    \addplot[color={rgb,1:red,0.2118;green,0.3922;blue,0.5451}, name path={8c8d5f98-6c0e-4762-bf89-a3ebcc442931}, draw opacity={1.0}, line width={2}, dashed]
        table[row sep={\\}]
        {
            \\
            3.0  -6.36  \\
            3.0  6.36  \\
        }
        ;
    \addplot[color={rgb,1:red,0.1804;green,0.5451;blue,0.3412}, name path={a9b18850-7f9f-437f-bf81-e3b3c6fa9c86}, draw opacity={1.0}, line width={2}, solid]
        table[row sep={\\}]
        {
            \\
            4.75  -6.36  \\
            4.75  6.36  \\
        }
        ;
     \addplot[color={rgb,1:red,0.0;green,0.0;blue,0.0}, name path={d91fbaa8-26a4-4dbe-8450-0f81039661ab}, only marks, draw opacity={1.0}, line width={0}, solid, mark=*,
    mark size={2.5 pt},
    mark repeat={1}, mark options={color={rgb,1:red,0.0;green,0.0;blue,0.0}, draw opacity={1.0}, fill={rgb,1:red,0.0;green,0.0;blue,0.0}, fill opacity={1.0}, line width={0.75}, rotate={0}, solid}]
        table[row sep={\\}]
        {
            \\
            0.0  0.0  \\
        }
        ;
    \addlegendentry {data\; }
    \addlegendentry {Var\;}
    \addlegendentry {MI\;}
    \addlegendentry {ME\;}
    \addlegendentry {PCov\;}
    \addlegendentry {WPCov\;}
\end{axis}
\end{tikzpicture}
\vspace{-2.0em}
\caption{%
    A visual comparison of the selection strategies PCov and WPCov, introduced in this paper, to the baselines Var, MI, and ME on a contrived data set.
    All strategies share the leftmost point (black circle) as an initial observation as well as the same exponentiated quadratic kernel and a zero mean prior.
    The vertical bars highlight which data point is sampled next for pivoting during the Cholesky decomposition.
    Depending on the strategy's objective, the correlation with other points, their proximity, the observation's value, or a combination of these quantities is decisive.
    See Section~\ref{sec:theory} and Appendix~\ref{asec:selection} for explanations and further information on each algorithm.
}
\label{fig:selection_strategies_overview}
\end{figure}

\begin{figure*}[t]
	\centering
    \hspace*{3.0em}
\begingroup%
\makeatletter%
\begin{pgfpicture}%
\pgfpathrectangle{\pgfpointorigin}{\pgfqpoint{5.181066in}{0.417746in}}%
\pgfusepath{use as bounding box, clip}%
\begin{pgfscope}%
\pgfsetbuttcap%
\pgfsetmiterjoin%
\definecolor{currentfill}{rgb}{1.000000,1.000000,1.000000}%
\pgfsetfillcolor{currentfill}%
\pgfsetlinewidth{0.000000pt}%
\definecolor{currentstroke}{rgb}{1.000000,1.000000,1.000000}%
\pgfsetstrokecolor{currentstroke}%
\pgfsetdash{}{0pt}%
\pgfpathmoveto{\pgfqpoint{0.000000in}{0.000000in}}%
\pgfpathlineto{\pgfqpoint{5.181066in}{0.000000in}}%
\pgfpathlineto{\pgfqpoint{5.181066in}{0.417746in}}%
\pgfpathlineto{\pgfqpoint{0.000000in}{0.417746in}}%
\pgfpathlineto{\pgfqpoint{0.000000in}{0.000000in}}%
\pgfpathclose%
\pgfusepath{fill}%
\end{pgfscope}%
\begin{pgfscope}%
\pgfsetrectcap%
\pgfsetroundjoin%
\pgfsetlinewidth{1.505625pt}%
\definecolor{currentstroke}{rgb}{1.000000,0.498039,0.054902}%
\pgfsetstrokecolor{currentstroke}%
\pgfsetdash{}{0pt}%
\pgfpathmoveto{\pgfqpoint{0.141667in}{0.219167in}}%
\pgfpathlineto{\pgfqpoint{0.280556in}{0.219167in}}%
\pgfpathlineto{\pgfqpoint{0.419444in}{0.219167in}}%
\pgfusepath{stroke}%
\end{pgfscope}%
\begin{pgfscope}%
\definecolor{textcolor}{rgb}{0.000000,0.000000,0.000000}%
\pgfsetstrokecolor{textcolor}%
\pgfsetfillcolor{textcolor}%
\pgftext[x=0.530556in,y=0.170556in,left,base]{\color{textcolor}\rmfamily\fontsize{10.000000}{12.000000}\selectfont Var}%
\end{pgfscope}%
\begin{pgfscope}%
\pgfsetrectcap%
\pgfsetroundjoin%
\pgfsetlinewidth{1.505625pt}%
\definecolor{currentstroke}{rgb}{0.549020,0.337255,0.294118}%
\pgfsetstrokecolor{currentstroke}%
\pgfsetdash{}{0pt}%
\pgfpathmoveto{\pgfqpoint{1.034774in}{0.219167in}}%
\pgfpathlineto{\pgfqpoint{1.173663in}{0.219167in}}%
\pgfpathlineto{\pgfqpoint{1.312552in}{0.219167in}}%
\pgfusepath{stroke}%
\end{pgfscope}%
\begin{pgfscope}%
\definecolor{textcolor}{rgb}{0.000000,0.000000,0.000000}%
\pgfsetstrokecolor{textcolor}%
\pgfsetfillcolor{textcolor}%
\pgftext[x=1.423663in,y=0.170556in,left,base]{\color{textcolor}\rmfamily\fontsize{10.000000}{12.000000}\selectfont Random}%
\end{pgfscope}%
\begin{pgfscope}%
\pgfsetrectcap%
\pgfsetroundjoin%
\pgfsetlinewidth{1.505625pt}%
\definecolor{currentstroke}{rgb}{0.580392,0.403922,0.741176}%
\pgfsetstrokecolor{currentstroke}%
\pgfsetdash{}{0pt}%
\pgfpathmoveto{\pgfqpoint{2.276459in}{0.219167in}}%
\pgfpathlineto{\pgfqpoint{2.415348in}{0.219167in}}%
\pgfpathlineto{\pgfqpoint{2.554237in}{0.219167in}}%
\pgfusepath{stroke}%
\end{pgfscope}%
\begin{pgfscope}%
\definecolor{textcolor}{rgb}{0.000000,0.000000,0.000000}%
\pgfsetstrokecolor{textcolor}%
\pgfsetfillcolor{textcolor}%
\pgftext[x=2.665348in,y=0.170556in,left,base]{\color{textcolor}\rmfamily\fontsize{10.000000}{12.000000}\selectfont ME}%
\end{pgfscope}%
\begin{pgfscope}%
\pgfsetrectcap%
\pgfsetroundjoin%
\pgfsetlinewidth{1.505625pt}%
\definecolor{currentstroke}{rgb}{0.121569,0.466667,0.705882}%
\pgfsetstrokecolor{currentstroke}%
\pgfsetdash{}{0pt}%
\pgfpathmoveto{\pgfqpoint{3.150713in}{0.219167in}}%
\pgfpathlineto{\pgfqpoint{3.289602in}{0.219167in}}%
\pgfpathlineto{\pgfqpoint{3.428491in}{0.219167in}}%
\pgfusepath{stroke}%
\end{pgfscope}%
\begin{pgfscope}%
\definecolor{textcolor}{rgb}{0.000000,0.000000,0.000000}%
\pgfsetstrokecolor{textcolor}%
\pgfsetfillcolor{textcolor}%
\pgftext[x=3.539602in,y=0.170556in,left,base]{\color{textcolor}\rmfamily\fontsize{10.000000}{12.000000}\selectfont PCov}%
\end{pgfscope}%
\begin{pgfscope}%
\pgfsetrectcap%
\pgfsetroundjoin%
\pgfsetlinewidth{1.505625pt}%
\definecolor{currentstroke}{rgb}{0.172549,0.627451,0.172549}%
\pgfsetstrokecolor{currentstroke}%
\pgfsetdash{}{0pt}%
\pgfpathmoveto{\pgfqpoint{4.165281in}{0.219167in}}%
\pgfpathlineto{\pgfqpoint{4.304169in}{0.219167in}}%
\pgfpathlineto{\pgfqpoint{4.443058in}{0.219167in}}%
\pgfusepath{stroke}%
\end{pgfscope}%
\begin{pgfscope}%
\definecolor{textcolor}{rgb}{0.000000,0.000000,0.000000}%
\pgfsetstrokecolor{textcolor}%
\pgfsetfillcolor{textcolor}%
\pgftext[x=4.554169in,y=0.170556in,left,base]{\color{textcolor}\rmfamily\fontsize{10.000000}{12.000000}\selectfont WPCov}%
\end{pgfscope}%
\end{pgfpicture}%
\makeatother%
\endgroup%
\\[-1em]
    \input{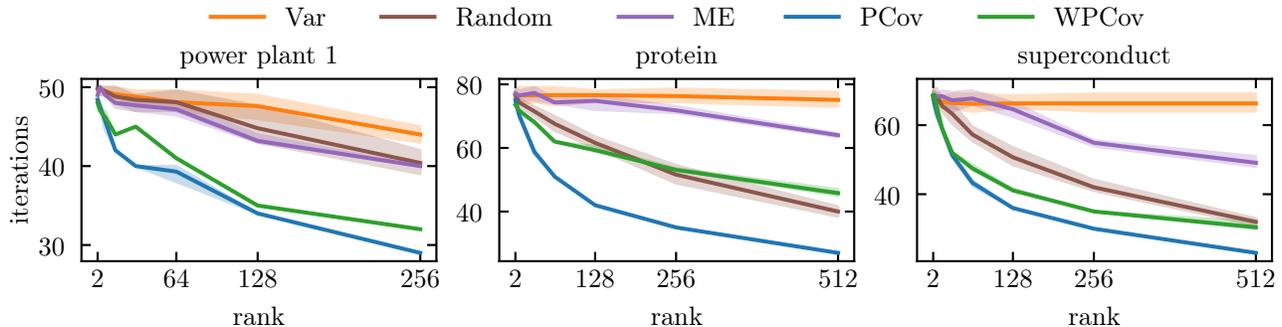}
    \vspace*{-2.0em}
	\caption{%
        Results of the preconditioned conjugate gradient experiments on three of the twelve data sets (columns).
        The performance of all methods is measured by the number of iterations until convergence.
        Lower values are better as the number of iterations is directly proportional to the wall clock runtime.
        The shaded areas mark the 90\% confidence interval originating from 10 repetitions with random permutations of the data set for all methods.
    }
	\label{fig:cg_convergence_results}
\end{figure*}

\section{Experiments}
\label{sec:experiments}
Every selection criterion considers a different quantity of interest, leading to a different set of included points which is indexed by $\I$. 
In this section, we benchmark the performance of the selection strategies introduced in Sections~\ref{sec:pcov}~and~\ref{sec:wpcov} based on metrics that are relevant to the respective downstream tasks: preconditioning linear systems and sparse GP regression.
The focus is placed on the evolution of these metrics as more points are included in $\I$.
Since these metrics are highly task-dependent, there can not be a single best algorithm.

In Section~\ref{sec:cg_convergence}, we use the selection strategies to build a low-rank Cholesky decomposition~\cite{cutajar2016preconditioning_kernel_matrices,liu2015pivotedcrossapproximation} that is employed as a preconditioner for conjugate gradients, for example to speed up GP regression~\cite{gardner2018gpytorch,mackay1997gp_cg}.
In Section~\ref{sec:gp_regression}, we show examples of sparse GP regression for different selection strategies and explore the respective impact on the trace, $\ell_2$-norm, and NLML~\eqref{eq:elbo}.

\textbf{Data Preprocessing.}
We chose twelve popular ML data sets curated by the University of California~\cite{Dua_ML_datasets}.
The training data was standardized to zero mean and unit variance in every dimension.
Moreover, we removed samples with missing values and randomly shuffled the remaining samples for every experiment.
In order to run the sparse GP regression experiments (Section~\ref{sec:gp_regression}) on a 6-core Intel Xeon CPU with 32 GB RAM, we truncated the data sets by randomly selecting 7000 samples.
For the CG preconditioner experiments (Section~\ref{sec:cg_convergence}) we employed a lazy implementation for matrix multiplications~\cite{CovarianceFunctionsRepo} which enabled us to use the complete data sets with up to 45730 samples.

\textbf{Model.}
Throughout the experiments, we used a zero mean GP prior together with an exponentiated quadratic kernel, ARD-transformed inputs, and additive homogeneous noise.
For every GP, the hyperparameters were optimized on a random subset of maximally 7000 data points, using evidence maximization~\cite{rasmussen2006gaussian}, and subsequently fixed during evaluation.

\textbf{Approximation of the MI Selection.}
As stated in Section~\ref{sec:pcov}, the PCov selection strategy is related to the MI, making it an interesting baseline.
Due to its high computational complexity of $\mathcal{O}(MN^4)$, we had to use an approximation of the MI~\cite{krause2008mutualinformation} specifically tuned for stationary kernels.
With this, data points locally influence each other only if they are sufficiently correlated, i.e., $|k_{ij}| / \sqrt{k_{ii}k_{jj}} > \varepsilon$ where $\varepsilon \in [0,1]$ is a relative tolerance threshold.
The quality of the selected points of the MI baseline generally improves with a lower threshold.
Due to the complex interaction between data set size, data input distribution, and kernel hyperparameters, it was prohibitively expensive to tune the threshold for every data set, so we fixed the value at $\varepsilon=0.5$ for all experiments.
The same idea of local interactions could also be used to speed up the other selection strategies, however, it is not necessary as they all just add at most a single MVP of computational cost to the Cholesky decomposition.
For the full data sets used in Section~\ref{sec:cg_convergence}, the approximated computation of the MI takes longer than solving the linear systems.
Thus, we replace this baseline with a selection strategy that samples the next pivoting point randomly.

\begin{figure*}[t]
	\centering
    \hspace*{3.0em}
\begingroup%
\makeatletter%
\begin{pgfpicture}%
\pgfpathrectangle{\pgfpointorigin}{\pgfqpoint{4.766840in}{0.417746in}}%
\pgfusepath{use as bounding box, clip}%
\begin{pgfscope}%
\pgfsetbuttcap%
\pgfsetmiterjoin%
\definecolor{currentfill}{rgb}{1.000000,1.000000,1.000000}%
\pgfsetfillcolor{currentfill}%
\pgfsetlinewidth{0.000000pt}%
\definecolor{currentstroke}{rgb}{1.000000,1.000000,1.000000}%
\pgfsetstrokecolor{currentstroke}%
\pgfsetdash{}{0pt}%
\pgfpathmoveto{\pgfqpoint{0.000000in}{0.000000in}}%
\pgfpathlineto{\pgfqpoint{4.766840in}{0.000000in}}%
\pgfpathlineto{\pgfqpoint{4.766840in}{0.417746in}}%
\pgfpathlineto{\pgfqpoint{0.000000in}{0.417746in}}%
\pgfpathlineto{\pgfqpoint{0.000000in}{0.000000in}}%
\pgfpathclose%
\pgfusepath{fill}%
\end{pgfscope}%
\begin{pgfscope}%
\pgfsetrectcap%
\pgfsetroundjoin%
\pgfsetlinewidth{1.505625pt}%
\definecolor{currentstroke}{rgb}{1.000000,0.498039,0.054902}%
\pgfsetstrokecolor{currentstroke}%
\pgfsetdash{}{0pt}%
\pgfpathmoveto{\pgfqpoint{0.141667in}{0.219167in}}%
\pgfpathlineto{\pgfqpoint{0.280556in}{0.219167in}}%
\pgfpathlineto{\pgfqpoint{0.419444in}{0.219167in}}%
\pgfusepath{stroke}%
\end{pgfscope}%
\begin{pgfscope}%
\definecolor{textcolor}{rgb}{0.000000,0.000000,0.000000}%
\pgfsetstrokecolor{textcolor}%
\pgfsetfillcolor{textcolor}%
\pgftext[x=0.530556in,y=0.170556in,left,base]{\color{textcolor}\rmfamily\fontsize{10.000000}{12.000000}\selectfont Var}%
\end{pgfscope}%
\begin{pgfscope}%
\pgfsetrectcap%
\pgfsetroundjoin%
\pgfsetlinewidth{1.505625pt}%
\definecolor{currentstroke}{rgb}{0.839216,0.152941,0.156863}%
\pgfsetstrokecolor{currentstroke}%
\pgfsetdash{}{0pt}%
\pgfpathmoveto{\pgfqpoint{1.034774in}{0.219167in}}%
\pgfpathlineto{\pgfqpoint{1.173663in}{0.219167in}}%
\pgfpathlineto{\pgfqpoint{1.312552in}{0.219167in}}%
\pgfusepath{stroke}%
\end{pgfscope}%
\begin{pgfscope}%
\definecolor{textcolor}{rgb}{0.000000,0.000000,0.000000}%
\pgfsetstrokecolor{textcolor}%
\pgfsetfillcolor{textcolor}%
\pgftext[x=1.423663in,y=0.170556in,left,base]{\color{textcolor}\rmfamily\fontsize{10.000000}{12.000000}\selectfont MI}%
\end{pgfscope}%
\begin{pgfscope}%
\pgfsetrectcap%
\pgfsetroundjoin%
\pgfsetlinewidth{1.505625pt}%
\definecolor{currentstroke}{rgb}{0.580392,0.403922,0.741176}%
\pgfsetstrokecolor{currentstroke}%
\pgfsetdash{}{0pt}%
\pgfpathmoveto{\pgfqpoint{1.862234in}{0.219167in}}%
\pgfpathlineto{\pgfqpoint{2.001123in}{0.219167in}}%
\pgfpathlineto{\pgfqpoint{2.140012in}{0.219167in}}%
\pgfusepath{stroke}%
\end{pgfscope}%
\begin{pgfscope}%
\definecolor{textcolor}{rgb}{0.000000,0.000000,0.000000}%
\pgfsetstrokecolor{textcolor}%
\pgfsetfillcolor{textcolor}%
\pgftext[x=2.251123in,y=0.170556in,left,base]{\color{textcolor}\rmfamily\fontsize{10.000000}{12.000000}\selectfont ME}%
\end{pgfscope}%
\begin{pgfscope}%
\pgfsetrectcap%
\pgfsetroundjoin%
\pgfsetlinewidth{1.505625pt}%
\definecolor{currentstroke}{rgb}{0.121569,0.466667,0.705882}%
\pgfsetstrokecolor{currentstroke}%
\pgfsetdash{}{0pt}%
\pgfpathmoveto{\pgfqpoint{2.736488in}{0.219167in}}%
\pgfpathlineto{\pgfqpoint{2.875377in}{0.219167in}}%
\pgfpathlineto{\pgfqpoint{3.014266in}{0.219167in}}%
\pgfusepath{stroke}%
\end{pgfscope}%
\begin{pgfscope}%
\definecolor{textcolor}{rgb}{0.000000,0.000000,0.000000}%
\pgfsetstrokecolor{textcolor}%
\pgfsetfillcolor{textcolor}%
\pgftext[x=3.125377in,y=0.170556in,left,base]{\color{textcolor}\rmfamily\fontsize{10.000000}{12.000000}\selectfont PCov}%
\end{pgfscope}%
\begin{pgfscope}%
\pgfsetrectcap%
\pgfsetroundjoin%
\pgfsetlinewidth{1.505625pt}%
\definecolor{currentstroke}{rgb}{0.172549,0.627451,0.172549}%
\pgfsetstrokecolor{currentstroke}%
\pgfsetdash{}{0pt}%
\pgfpathmoveto{\pgfqpoint{3.751055in}{0.219167in}}%
\pgfpathlineto{\pgfqpoint{3.889944in}{0.219167in}}%
\pgfpathlineto{\pgfqpoint{4.028833in}{0.219167in}}%
\pgfusepath{stroke}%
\end{pgfscope}%
\begin{pgfscope}%
\definecolor{textcolor}{rgb}{0.000000,0.000000,0.000000}%
\pgfsetstrokecolor{textcolor}%
\pgfsetfillcolor{textcolor}%
\pgftext[x=4.139944in,y=0.170556in,left,base]{\color{textcolor}\rmfamily\fontsize{10.000000}{12.000000}\selectfont WPCov}%
\end{pgfscope}%
\end{pgfpicture}%
\makeatother%
\endgroup%
\\[-1em]
    \input{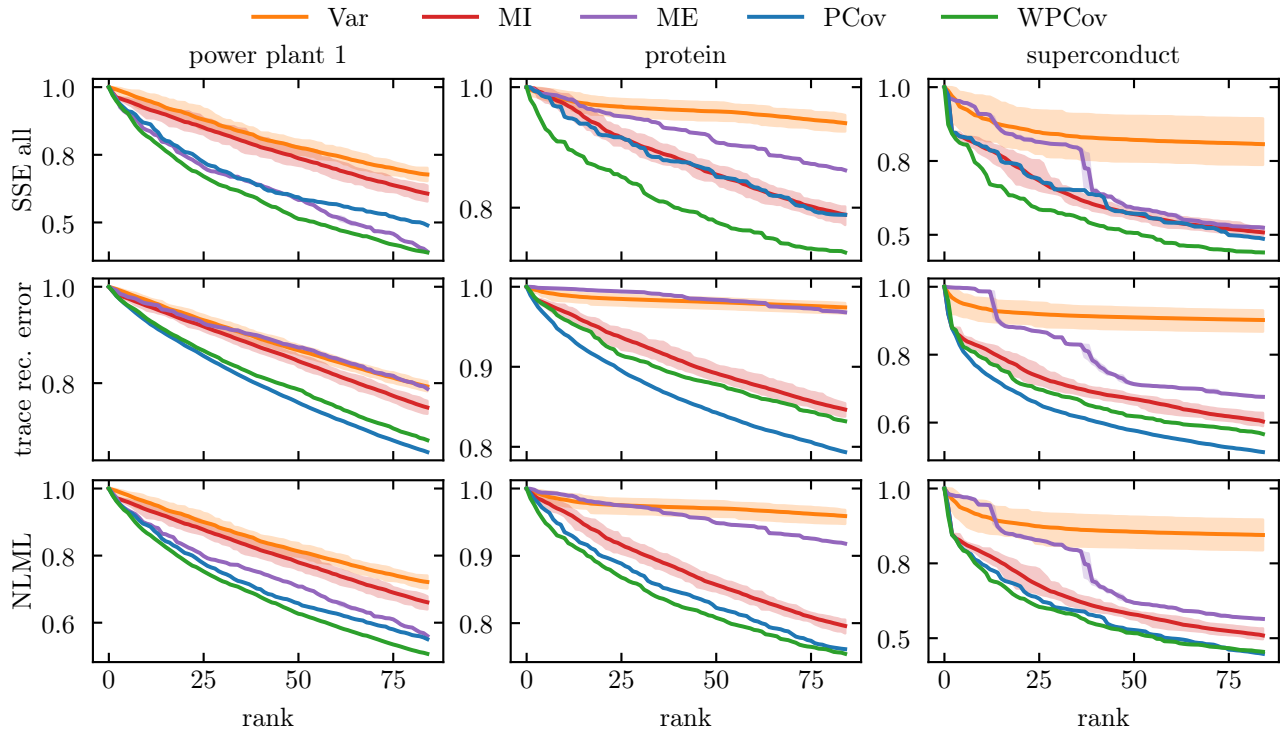}
    \vspace*{-2.0em}
	\caption{%
        Results of the sparse GP regression experiments on three of the twelve data sets (columns).
        The performance of all methods is measured on the metrics (rows) described in Section~\ref{sec:gp_regression}.
        Lower values are better for all metrics, which were normalized to facilitate comparison among the data sets.
        The shaded areas mark the 90\% confidence interval originating from 20 repetitions with random permutations of the data set for all methods.
    }
	\label{fig:results_gp_regression}
\end{figure*}
\subsection{Preconditioned Conjugate Gradients}
\label{sec:cg_convergence}
One efficient way to alleviate the computationally expensive task of solving a large linear system $G\alpha = y$ where $G=K_{X X}+\sigma^2 I$, is to use an iterative linear solver such as conjugate gradients (CG)~\cite{hestenes1952cg} that sequentially approaches the true solution.
It is often advised, and sometimes necessary, to use a preconditioner to speed up and robustify the solver's convergence.
A popular preconditioner is based on a partial pivoted Cholesky decomposition~\cite{gardner2018gpytorch,harbrecht2012pivoted_cholesky} which corresponds to a FITC-type approximation, as described in Appendix~\ref{asec:preconditioner}.
\citeauthor{cutajar2016preconditioning_kernel_matrices}~\cite{cutajar2016preconditioning_kernel_matrices} established that low-rank approximations of this kind are generic and efficient preconditioners for GP inference.

\textbf{Experiment Setup.}
To quantify the selection strategies' influence on a preconditioner's performance, we executed an experiment where the same linear system is solved by preconditioned CG until a relative error tolerance of $10^{-4}$ is reached.
For this experiment, we implemented a new array type that better leverages the inherent low-rank structure of the Cholesky decomposition for reduced memory allocation and faster matrix inversion compared to the compressed-sparse-column format commonly used.
Further details on our array type can be found in Appendix~\ref{asec:preconditioner}.

\textbf{Chosen Metric.}
A selection strategy is considered better when its associated preconditioner allows the CG-solver to converge in fewer iterations. 
We evaluate at the ranks $2^{1:R}$ with $R = \lceil \log_2( \sqrt{N} ) \rceil$ + 1.

\textbf{Results.}
We found that our proposed selection strategies, especially PCov, lead to preconditioners that speed up the CG solver in terms of the number of iterations until convergence (Figure~\ref{fig:cg_convergence_results}) which directly translates to savings in the wall clock time (Figure~\ref{afig:cg_convergence_results_all} in Appendix~\ref{asec:gp_regression}).
The results are particularly important because of the synergy with modern machine learning frameworks that utilize the parallel capacity of the GPU together with CG for training~\cite{gardner2018gpytorch,artemev2021cg_bounds,davies2015effective,meanti2020kcg}.

\subsection{Sparse Gaussian Process Regression}
\label{sec:gp_regression}
To ascertain whether the new pivoting strategies also benefit sparse GP regression applications, we evaluated every strategy based on the samples it selected from the data set.

\textbf{Chosen Metrics.}
The goal of the downstream task is to make accurate predictions $\hat{y}$, which can be measured by the sum of squared errors $\sum_{i=1}^{N} (y_i - \hat{y}_i)^2$.
Minimizing this metric results in the least-squares estimator ${ \alpha_{\I} = (K_{\I \perm{X}}K_{\perm{X} \I})^{-1}K_{\I \perm{X}} \perm{y} }$ as new points are included in $\I$.
The motivation behind PCov and (partially) its weighted variant WPCov is to greedily add a point to $\I$ in order to reduce the variance of the remaining points in $\R$. 
To this end, we chose to evaluate the trace of the reconstruction error $\tr(G_{\perm{X}\perm{X}} - L_{\perm{X}\I} L_{\perm{X}\I}\trans)$ which measures the remaining total variance. 
Training a sparse GP with variational inference, which is out of the scope of this paper, is done by maximizing the ELBO which is equivalent to minimizing the NLML~\eqref{eq:elbo}~\cite{burt19convergence}.
Thereby, pivoting strategies that yield approximations of the kernel matrix with a lower NLML faster, i.e., at lower ranks, are beneficial for training sparse GPs in terms of number of samples.
Whether it translates to faster training is highly dependent on the implementation and computational resources and is not explored further.
We evaluate at the ranks $1 : \lceil\sqrt{N}\rceil$ where $N$ is the (truncated) data set size.

\textbf{Results.}
Our results (Figure~\ref{fig:results_gp_regression}) indicate that PCov and WPCov are favorable trade-offs in terms of performance and computational cost. 
Particularly the evolution of the reconstruction error's trace shows that both PCov and WPCov can efficiently reduce the uncertainty.
A complementary interpretation is that PCov and WPCov reduce the spectral range of the Gram matrix faster since the trace of a matrix is equal to the sum of its eigenvalues.
For kernels with rapidly decaying spectrum like the exponentiated quadratic~\cite{wenger2022gphyperparameter}, it would likely lead to a reduced condition number which supports the claim that the new selection criteria yield good preconditioners for iterative linear solvers (Section~\ref{sec:cg_convergence} and Appendix~\ref{asec:preconditioner}). 
The sum of squared errors (SSE) results also show that WPCov can efficiently identify data points that lower the modeling error significantly, often outperforming the ME.
The SSE as well as the trace are closely related to the data-fitting and penalty terms of the NLML~\eqref{eq:elbo}, partially explaining the good performance of PCov and WPCov in this metric.
Moreover, PCoV and WCPov are invariant w.r.t. the order of the data, hence the confidence interval is zero and any deviations are due to numerical imprecision.
Additional experiments and the complete evaluation can be found in Appendix~\ref{asec:gp_regression}.

\section{Additional Related Work}
The pivoted Cholesky decomposition has lately received renewed attention in the GP community.
\citeauthor{burt19convergence}~\cite{burt19convergence} showed, among other things, that it can serve as a better initialization for continuous optimization of inducing point methods~\cite{chang2023piv_ind} compared to traditional K-means initialization~\cite{hensman2013bigdata}.
Furthermore, it is possible to avoid the continuous optimization of the inducing points in~\eqref{eq:elbo} by selecting $X_{\I}$ according to~\eqref{eq:variance} and only optimizing the kernel hyperparameters.

\citeauthor{schaefer2021sparse_cholesky}~\cite{schaefer2021sparse_cholesky} developed a form of incomplete Cholesky decomposition with particular application to kernel matrices that arise in partial differential equations and GPs.
The authors choose the next pivot point as the one that is the furthest away from the previously seen points \emph{and} the border of the input space.
In the densely sampled input space that the authors considered, their selection would be similar to PCov (Section~\ref{sec:pcov}).
In a more general setting, where data is sparse or clustered and the suitable bounds of the data are unknown, the two approaches differ since maximizing distance to the bounds and active set does not imply that the point is close to many other points, which is the criterion for PCov~\eqref{eq:pcov}.
%
%

When the exponentiated quadratic kernel is used, then PCov (Section~\ref{sec:pcov}) can be seen as a special case of the information density introduced in~\cite{settles2008alanalysis}.
The acquisition function $\alpha(x_j) = \phi(x_j) \sum_i \operatorname{sim}(x_j,x_i)$ weights a statistical quantity $\phi(x)$, like the variance, by a similarity measure to all other points, where the distance can be L2, cosine, or a squared exponential. The correspondence is no longer valid for other stationary kernels.

\citeauthor{cutajar2016preconditioning_kernel_matrices}~\cite{cutajar2016preconditioning_kernel_matrices} showed how GP inference and hyperparameter training can be achieved by only using iterative methods.
The authors further explored the effect of preconditioned CG with different matrix approximations and found that preconditioners constructed as a FITC or \nystrom approximation with a random subset of the data can significantly reduce the required number of iterations.
\citeauthor{wenger2022gphyperparameter}~\cite{wenger2022gphyperparameter} took the idea of preconditioning one step further and showed how a partial Cholesky decomposition with standard pivoting could be used to reduce the variance of stochastic trace estimators with applications to GP hyperparameter training\cite{artemev2021cg_bounds}.
Our work can be seen as a natural progression to improve these ideas by investing more in generating a better \nystrom approximation in order to speed up the downstream tasks of preconditioned CG.

\section{Conclusion}
In this work, we combined ideas from active learning with the pivoted Cholesky decomposition, focusing on the application of GP inference.
By doing so we made popular greedy selection strategies from the GP literature available as strategies for the pivoted Cholesky decomposition.
We additionally introduced two selection strategies (PCov and WPCov) that exhibit great potential as a preconditioner for CG as well as for inducing point selection during sparse GP regression.

\subsection{Limitations}
The reasoning behind PCov (Section~\ref{sec:pcov}) assumes that all the elements of the matrix are positive.
This is true for Gram matrices produced by most stationary covariance functions but is not a defining property of SPD matrices in general.
However, the flexibility of WPCov (Section~\ref{sec:wpcov}) to use an arbitrary vector along with its performance suggests that this limitation can be mitigated by an appropriately chosen, task-dependent, projection vector.

For the pivoting selection, we have only considered the information of a single vector containing a summary statistic.
By incorporating multiple sources~\cite{seeger2003fast} one could tune the selection to specialized use cases.

An important limitation is that the set of included data points $X_{\I}$ is obtained from a greedy selection strategy, hence in general does not result in the optimal set of points.
Jointly-optimized selections are unfortunately intractable for most problems.

Moreover, we want to mention that all experiments were conducted using real-valued matrices, leaving an analysis for complex Hermitian matrices as future work.

\subsection{Outlook}
A practical feature that has not been explored yet would be to automatically terminate the pivoted Cholesky routine based on the evolution of a task-dependent summary statistic.
This would for example offer a natural estimate for the minimum rank of a preconditioner for linear systems (Section~\ref{sec:cg_convergence}) without increasing the computational cost.

The active selection of points used by the pivoted Cholesky decomposition (Section~\ref{sec:theory}) resembles the policy of a learning agent.
This idea lies at the core of the field of probabilistic numerics where direct solution methods so far have received little attention in favor of iterative linear solvers~\cite[][Ch.~5]{hennig2022pn}.
Exploring this connection is a promising avenue for future research.

Finally, it would also be interesting to see how the proposed selection strategies apply to other forms of inference such as multi-output GPs~\cite{alvarez2012mo_kernels}, potentially with a special structure~\cite{gessner21gp_derivative, ament2022gp_bo}, spectral approximations~\cite{rahimi2007rff}, or classification tasks~\cite[][Ch.~3]{rasmussen2006gaussian}.



\newpage
\printbibliography 




\newpage
\appendix
\section{Preconditioning Linear Systems}
\label{asec:preconditioner}
There are two main approaches to solving a linear system of equations $G\alpha = y$ where $G$ is an SPD matrix.
A direct solution method first performs an expensive matrix decomposition of $G$, such as the Cholesky decomposition, see Section~\ref{sec:cholesky} or~\cite{golub2013matrix}.
The resulting factors, i.e., the lower or upper triangular, are then used to solve the linear system efficiently.

\textbf{Using the Cholesky to generate preconditioners for CG.}
Another option is to use an iterative solver such as conjugate gradients~\cite{hestenes1952cg} which can speed up the solution if $G$ has a clustered spectrum or a limited precision is required in the solution.
This approach has the benefit that CG does not require $G$ to be stored explicitly which can be prohibitive for kernel learning with large data sets.
Instead, it only requires access to matrix-vector multiplications of $G$ which in the case of kernel functions can be efficiently implemented to run on a GPU~\cite{gardner2018gpytorch}.
The convergence rate of CG depends on the square root of the condition number 
\begin{equation}
  \kappa (G) = \frac{\lambda_{\text{max}} (G)}{\lambda_{\text{min}} (G)},
\end{equation}
which is the ratio of the largest and smallest eigenvalue of $G$.
It is possible and often necessary to improve the convergence rate of CG by using a preconditioner.
$LL\trans \approx G$ then one can instead solve the equivalent linear system with the matrix $\perm{G} = L^{-1} G L \mtrans$.
Preferably $\kappa(\perm{G}) \ll \kappa(G)$ to ensure rapid convergence towards the solution.
There are other ways to phrase the preconditioning but it is important that it is cheap to apply and low-rank matrices or partial Cholesky decompositions are popular generic options.

\textbf{Introducing a Task-Specific Data Structure.}
Our new data structure is tailored to triangular matrices with a low-rank structure~\cite{liu2015pivotedcrossapproximation}.
Without loss of generality, we use the indices $\I={1,\dots,M}$ and $\R={M+1,\dots,N}$ in order to simplify the derivation of the preconditioner.
We define the diagonal matrix
\begin{align}
    \perm{D}^2_{\R} &= 
    \diag \left( K_{XX} - L_{X \I}L_{\I X} \right) \in \R^{N\times N} \notag\\
    &= \begin{bmatrix}
    0_{M\times M} & \multicolumn{2}{c}{\cdots} & 0_{M \times 1} \\
    \multirow{2}{*}{\vdots}  & \perm{D}^2_{\R,M+1,M+1} & & 0_{\text{triang.}} \\
     & & \ddots & \\
    0_{1 \times M} & 0_{\text{triang.}} & & \perm{D}^2_{\R,N,N}
    \end{bmatrix}
\end{align}
which is zero for the first $M$ rows as well as columns, and has only entries on the diagonal for the remaining $(N-M) \times (N-M)$ block.
We further define
\begin{equation}
    \perm{L} = \perm{D}_{\R} + L_{X\I}E_{\I}\trans,
\end{equation}
which will be our approximated lower-triangular Cholesky factor~\cite{liu2015pivotedcrossapproximation}. Reconstructing the approximated matrix results in 
\begin{align}
    \perm{L}\perm{L}\trans 
    &= \perm{D}_{\R}^2 + L_{X\I}\underbrace{ E_{\I}\trans \perm{D}_{\R} }_{0} + \underbrace{\perm{D}_{\R}E_{\I}L_{\I X}}_{0} \notag\\
    &\phantomrel{=}  + L_{X\I}\underbrace{E_{\I}\trans E_{\I}}_{I_M}L_{\I X} \notag\\
    &= \perm{D}_{\R}^2 +  L_{X\I}L_{\I X} \label{aeq:reconstructed_LLtrans}.
\end{align}
It is worth noting that the last expression above is equivalent to the fully independent training conditional (FITC)
\begin{equation}
    \est{K}_{\text{FITC}} =
    \diag(K_{XX} - \est{K}_{XX}) + \est{K}_{XX}.
\label{aeq:fitc}
\end{equation}
where $\est{K}_{XX}$ is the \nystrom approximation~\eqref{eq:nystrom}.
This becomes apparent by expanding \eqref{aeq:reconstructed_LLtrans} into
\begin{align}
    \perm{D}_{\R}^2 +  L_{X\I}L_{\I X}
    &= \diag (K_{X X} - K_{X \I}(K_{\I\I})^{-1} K_{\I X} ) \notag\\
    &\phantomrel{=} + K_{X \I}(K_{\I\I})^{-1} K_{\I X}.
\end{align}


\begin{figure}[t]
    \centering
    \includegraphics[width=0.7\columnwidth]{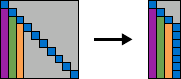}
    \caption{%
    The new data structure (right) proposed for sparse triangular matrices (left) that for example occur during a pivoted Cholesky decomposition.
    In this illustrative example, the decomposition $K=LL\trans$ is currently at rank 3, thus the first three columns of $L$ are nonzero below the diagonal (colored).
    These columns are stored fully in memory, as usual for dense arrays.
    The numbers on the diagonal of $L$ (blue) are generally nonzero.
    All remaining entries (gray) are zero, hence, do not hold any information in the context of a Cholesky decomposition.
    }
    \label{afig:data_structure}
\end{figure}

\begin{figure*}[t]
  \centering
  \begin{subfigure}[t]{\columnwidth}
      \includegraphics[width=\columnwidth]{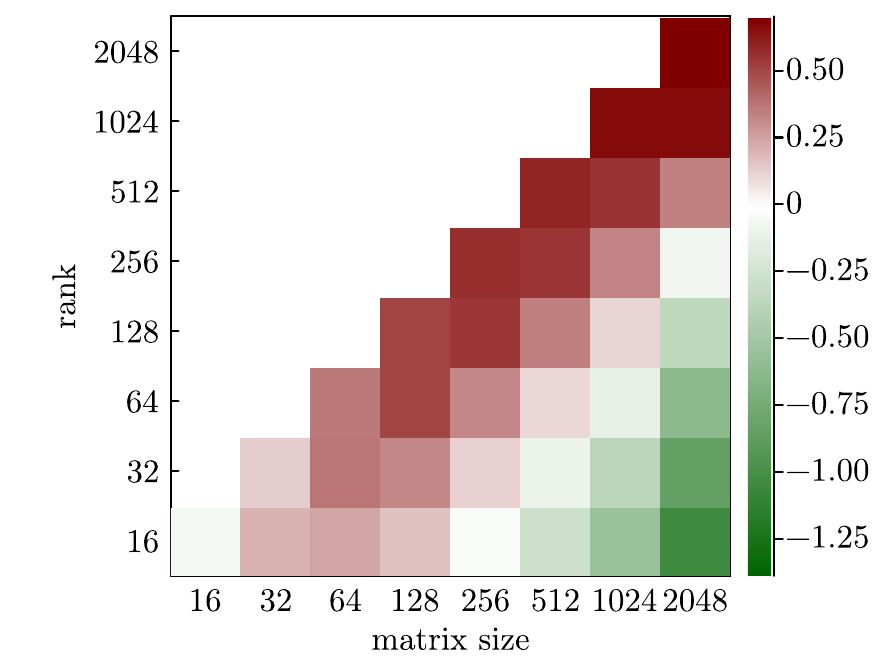}
      \caption{implementation from SparseArrays.jl}
  \end{subfigure}
  \hfill
  \begin{subfigure}[t]{\columnwidth}
      \includegraphics[width=\columnwidth]{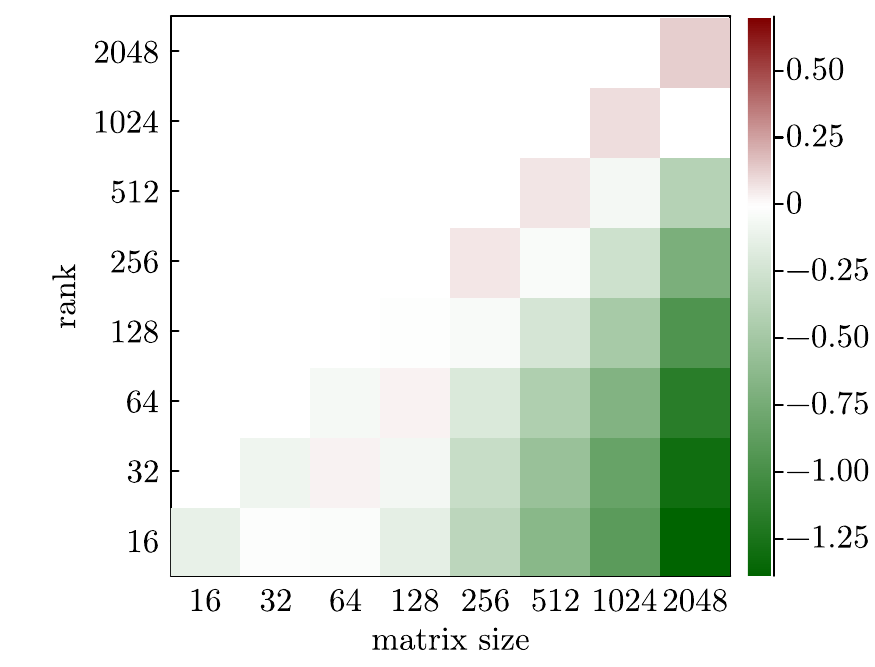}
      \caption{our implementation}
  \end{subfigure}
  \caption{%
  Benchmarking of the (relative) median time to solve a linear system of equations up to a given rank using the in-place left-hand side division.
  The time measurements are normalized by the results of Julia's standard implementation for dense arrays \emph{and} then $\log_{10}$-transformed, i.e., lower values are better.
  }
  \label{afig:benchmarking_implementations}
\end{figure*}

\textbf{Implementation in Julia.}
In contrast to other (sparse) array implementations, we store the diagonal for entries greater than the current rank of the approximation as a separate column (Figure~\ref{afig:data_structure}).
This way, we can omit storing the vast majority of zeros, while at the same time having a dense array that is indexed faster than sparse arrays from the Julia SparseArrays.jl~\cite{SparseArraysRepo} package.
To visualize the benefits of our custom implementation, we benchmarked the two implementations on the exemplary task of solving a linear system of equations which is equivalent to applying the preconditioner in the experiments of Section~\ref{sec:cg_convergence}.
The results (Figure~\ref{afig:benchmarking_implementations}) show that our data structure significantly accelerates solving linear systems for higher ranks, especially with increasing matrix size.
Note the log-scaling of the values.
This operation is decisive for the experiments in Section~\ref{sec:cg_convergence}.
Moreover, our new implementation has a lower memory footprint than the baselines.

\section{Additional Analysis of the Selection Strategies}
\label{asec:selection}

Before we identify and derive some of the popular selection strategies presented in the paper, we extend the notation from Section~\ref{sec:background} and discuss relevant properties of SPD matrices.

\textbf{Extended notation.}
We define the new sets
\begin{equation}
    \Ip = \I \cup j, \quad \Rm = \R \setminus j
\end{equation}
for any point $j\in \R$, and $\Im$ is the included set at the previous iteration.

\textbf{Schur Determinant.}
One matrix identity that will play a prominent role in the derivations involves the determinant of the Schur complement~\cite[][A.5.5]{boyd2004convex}
\begin{align}
    |K_{X X}|
    &= |K_{\perm{X} \perm{X}}| \notag\\ 
    &= |K_{\I \I}| \cdot |K_{\R \R} - K_{\R \I }(K_{\I \I})^{-1} K_{\I \R} |.\label{aeq:schur_determinant}
\end{align}
Just as in the main paper, all random variables are assumed to be Gaussian.

Finally, in an abuse of notation, we will refer to the entropy of a Gaussian process evaluated at all points $X$ as $\H(X)$ instead of $\H(f_X)$.

\textbf{Entropy of Gaussian Random Variable.}
The reason it makes an appearance is because the entropy of a multivariate normal-distributed random variable with covariance $K_{XX}$ is
\begin{align}
    \H (X)
    &= \frac{1}{2}\log \left( (2\pi e)^N \cdot |K_{XX}| \right) \notag\\ 
    &= \frac{N}{2} \left(1 + \log(2\pi) \right) + \frac{1}{2}\log \left(|K_{XX}| \right) \notag\\
    &\propto \log |K_{XX}|,\label{aeq:entropy}    
\end{align}
which will be important for the D-optimal selection (Section~\ref{asec:d-optimal}) and the mutual information (Section~\ref{asec:mi}). 

\subsection{D-optimal Selection}
\label{asec:d-optimal}
In D-optimal design, the goal is to maximize the determinant of the covariance matrix.
In the GP setting that is equivalent to finding an $i$ such that $|K_{\Ip \Ip}|$ is maximized.
This amounts to an entropy maximization algorithm by combining the definition of the entropy \eqref{aeq:entropy} with $\argmax \log |K_{\Ip \Ip}| = \argmax |K_{\Ip \Ip}|$.
The last property is due to the monotonic nature of the logarithm so both formulations share the same optimizer.
Hence, the goal to find $j$ to maximize $|K_{\Ip \Ip}|$ which we expand using \eqref{aeq:schur_determinant} into
\begin{equation}
    |K_{\Ip \Ip}| = |K_{\I \I}| \cdot |K_{jj} -  K_{j \I} (K_{\I \I})^{-1}K_{\I j}|.
    \label{aeq:variance_det}
\end{equation}
The determinant of $K_{\Ip \Ip}$ is maximized when the last term, which is the expression for $\Var ( j\mid \I)$, is maximized.
In other words, D-optimal design aims to find the point with maximal variance in $\R$ when conditioning on $\I$. 
For a stationary kernel, this will be the point $i$ that is furthest away from $\I$\footnote{It is measured by the underlying metric of the covariance function which does not have to be isotropic.}. 
This highlights the connection between variance maximization, D-optimal design, and the pivoted Cholesky decomposition.

\paragraph{The Variance Criterion Greedily Maximizes $|K_{\I\I}|$.}
The main text claims \enquote{The entropy selection employed by the standard pivoted Cholesky routine in~\eqref{eq:variance} is greedily selecting data points in order to maximize  $|K_{\I \I}|$}.

This can be seen as follows, by denoting $j$ as the new point to be added from $\R$ to $\I$ as in~\eqref{eq:variance}:
\begin{align}
    &  \argmax_{j\in\R} |K_{\Ip \Ip}| \notag\\
    &= \argmax_{j\in\R} |K_{\I\I}| \cdot |K_{jj} - K_{j\I}K_{\I\I}^{-1}K_{\I j}| \notag\\
    &= \argmax_{j\in\R} |K_{jj} - K_{j\I}K_{\I\I}^{-1}K_{\I j}| \notag\\
    &= \argmax_{j\in R} \diag(K_{\R\R} - K_{\R\I}K_{\I\I}^{-1}K_{\I\R})_{j}
\end{align}
where we could use the factorization of determinants for block matrices as $K_{\I\I}$ is invertible. 

\subsection{Mutual Information (MI)}
\label{asec:mi}
The arguably right thing to do is to find points that are most informative about the rest of the points, i.e., which point shares the most mutual information with the rest of the points. The mutual information is symmetric ($j$ reveals as much about $\Rm$ as the converse on average) and defined as
\begin{align}
\label{aeq:mi}
    &\MI\left( j,\Rm \right) =  \\
    &=\H(\R) - \H(\Rm \mid j) - \H(j\mid \Rm) \notag \\
    &= \H(j) - \H(j\mid \Rm) \notag \\
    &= \log \big( |K_{jj}|/|K_{jj} - K_{j \Rm}(K_{\Rm \Rm})^{-1}K_{\Rm j}| \big) \notag\\
    &= \H(\Rm) - \H(\Rm \mid j) \notag\\
    &= \log \big( |K_{\Rm \Rm}|/|K_{\Rm \Rm} - K_{\Rm j}(K_{j j})^{-1}K_{ j \Rm}| \big) \notag.
\end{align}
Evaluating this criterion for $N$ points has complexity $\mathcal{O}(N^4)$ because every point requires the inverse of a $(N-1)$-dimensional square matrix, and hence the cost to select $M$ points is $\mathcal{O}(MN^4)$.
By utilizing the local properties of a stationary kernel one can get a good approximation by only considering the covariance of nearby points.
If instead every point shares covariance with $Q$ other points then the cost decreases to $\mathcal{O}(NQ^3M)$ which is linear in $N$ and by tweaking $Q$ one can change the accuracy of the prediction.

\subsection{A-optimal Selection}
\label{asec:a-optimal}
An A-optimal selection seeks to minimize the trace of $K_{\R \R}$\footnote{$\tr (K_{\R \R}) = \sum_{j \in \R} K_{jj}$}. 
Selecting a point $x_j$, i.e., removing it from $\R$, will then result in a new trace
\begin{align}
    \tr (K_{\Rm \Rm})
    &=\tr(K_{\R \R} - K_{\R j} (K_{j j})^{-1}K_{j \R}) \notag\\
    &= \tr(K_{\R \R}) - \frac{\norm{ K_{\R j} }^2}{K_{j j}}.
 \end{align}
The goal of A-optimal design is then to maximize the second term which can be evaluated for every point at every iteration or recursively phrased by performing one MVP per iteration.
The overall cost of iteratively selecting $M$ points out of $N$ is $\mathcal{O}(MN^2)$.
\begin{align}
    \label{aeq:a_opt_mi}
    &\MI\left( j,\Rm \right) = \H(\R) - \H(\Rm \mid j) - \H(j\mid \Rm) \notag \\
    &\geq \H(\R) - \H(\Rm \mid j) - \H(j) \notag \\
    &\geq \H(\R) - \H^{+}(\Rm \mid j) - \H(j) \notag \\
    &\geq \H(\R) - \sum\limits_{i \in \Rm} \log \left(k_{ii} - \frac{k_{ij}^2}{k_{jj}}\right) - \log(k_{jj}) \notag \\
    &\geq \H(\R) - \sum\limits_{i \in \Rm} \left(k_{ii} - \frac{k_{ij}^2}{k_{jj}}\right) - k_{jj} \notag \\
\end{align}

Generally, this selection strategy would favor points with a high squared covariance with the remaining points.
For stationary kernels, these would be points that are close to many other points. 
However, the multiplicative scaling of $K_{jj}^{-1}$, which can be arbitrarily small for a point $j$ close to $\I$, can significantly alter this behavior.
For a stationary kernel, it means that a point $j$ close to $\I$, i.e., $\Var(j \mid \I)$ is small, is prioritized over \enquote{central} points, which can hinder exploration.






\subsection{Projected Covariance (PCov)}
\label{asec:pcov_implementation_details}
The newly introduced PCov selection strategy (Section~\ref{sec:pcov}) chooses the next point based on the value of $(K_{\perm{X}\perm{X}} - K_{\perm{X}\
}(K_{\I \I})^{-1}K_{\I\perm{X}} )\one_N$.
PCov shares several properties with a greedy A-optimal selection (Appendix~\ref{asec:a-optimal}) in that it will also identify points that share a high covariance, which for stationary kernels means that they are \enquote{central}. 
However, there are two differences that set the selection strategies apart.
First, A-optimal squares every covariance which means that a point that has a high covariance with a few points is preferred over a point that has a low covariance with many points. 
Second, and more importantly, PCov does not scale the sum by the inverse of the marginal variance which can be arbitrarily small, which for stationary kernels means that points close to $\I$ can be a very strong candidate for selection.
This in turn hampers the exploration but will lead to a lower trace in the end.

Moreover, in the case of stationary covariance functions, we can see PCov as a first-order approximation to the mutual information.

In the following sections, we provide detailed information as well as explanations on the update of PCov's statistic of interest (Section~\ref{asec:pcov_implementation_details_update}), and PCov's memory allocation (Section~\ref{asec:pcov_implementation_details_memory}).
Finally, we draw a short comparison to the standard implementations\footnote{\hyperlink{https://netlib.org/lapack/explore-html/da/dba/group__double_o_t_h_e_rcomputational_ga31cdc13a7f4ad687f4aefebff870e1cc.html}{dpstrf} in LAPACK and \hyperlink{https://www.netlib.org/linpack/dchdc.f}{dchdc} in LINPACK.} of the pivoted Cholesky decomposition (Section~\ref{asec:pcov_implementation_details_comparison}).

\subsubsection{Efficient Implementation of the PCov Update}
\label{asec:pcov_implementation_details_update}
In Section~\ref{sec:pcov} we claim that it is possible to speed up the computation of 
\begin{equation}
    s_{\perm{X}}^{(m+1)} = K_{\perm{X}\I}(K_{\I\I})^{-1}K_{\I \perm{X}} \one_{N}    
\end{equation}
from $\mathcal{O}(MN)$ to $\mathcal{O}(N)$, which we here outline in the three following steps.

\textbf{Step 1.}
We rewrite the update in terms of the Cholesky factor at iteration $m$ as 
\begin{align}
  s_{\perm{X}}^{(m+1)}
  &= K_{\perm{X}\I}(K_{\I\I})^{-1}K_{\I \perm{X}} \one_{N}   \notag\\
  &= L_{\perm{X}\I}L_{\I\I}\trans(L_{\I\I} L_{\I\I}\trans)^{-1} K_{\I \perm{X}}\one_{N}   \notag\\
  &= L_{\perm{X} \I}\underbrace{(L_{\I\I})^{-1}s^{*}_{\I}}_{z^{(m)}} \notag  \\
  &= L_{\perm{X} \I} z^{(m)}. \label{apcov:s_update}
\end{align}

\textbf{Step 2.}
We introduce the intermediate vector
\begin{equation}
    z^{(m)}=(L_{\I\I})^{-1}s^{*}_{\I} \in \Re^{m}
\end{equation}
which contains the solution to a linear system of equations with a lower-triangular matrix.
Because the matrix is lower-triangular, we can use the standard forward substitution\cite{golub2013matrix}[Ch. 3.1.1] to cache computations from previous iterations.
This means that if the solution at iteration $m-1$ is already computed and stored in
\begin{equation}
    z^{(m-1)} = (L_{\Im \Im})^{-1}s^{*}_{\Im}
\end{equation}
then the solution at iteration $m$ leaves the first $m-1$ indices unchanged and only requires an update to the $m$-th element $z^{(m)}_{m}$
\begin{align}
	z^{(m)} &=(L_{\I\I})^{-1}s^{*}_{\I} \notag\\
    &=
    \begin{bmatrix}
    z^{(m-1)}\\
    \left(s^{*}_{m} - \sum\limits_{j=1}^{m-1}L_{mj}z^{(m-1)}_j\right)/L_{mm}
    \end{bmatrix}.
\end{align}

\textbf{Step 3.} We use the fact that the indices ${[1, \dots, m-1]}$ of $z^{(m)}$ are not changed at iteration $m$ to get the following iterative/recursive update from \eqref{apcov:s_update}
\begin{align}
	s^{(m+1)}_{\perm{X}} 
    &= \sum\limits_{j=1}^{m} L_{\perm{X}j}z^{(m)}_{j} \notag\\  
    &= \sum\limits_{j=1}^{m-1} L_{\perm{X}j}z^{(m-1)}_{j} + L_{\perm{X}m}z^{(m)}_{m} \notag\\
    &=s^{(m)}_{\perm{X}} + L_{\perm{X}m}z^{(m)}_{m},
\end{align}
 which is just a vector addition of complexity $\mathcal{O}(N)$.

\subsubsection{Memory Allocation}
\label{asec:pcov_implementation_details_memory}
It is possible to circumvent the memory allocation of $z^{(m)}$ by instead overwriting the first $m$ elements of $s^{(m)}_{\perm{X}}$ to contain $z^{(m)}$. The full update is then a 2-step process consisting of an $\mathcal{O}(m)$ operation followed by an $\mathcal{O}(N-m)$. 
\begin{align}
    s^{(m)}_m &= \left(s^{*}_{m} - \sum\limits_{j=1}^{m-1}L_{mj}s^{(m)}_j\right)/L_{mm}\\
    s^{(m+1)}_{\R}  &=s^{(m)}_{\R } + L_{\R m}s^{(m)}_{m}
\end{align}

After iteration $m$ we then have the following information stored in $s^{(m)}_{\perm{X}}$
\begin{equation}
    s^{(m)}_{\perm{X}} = 
    \begin{bmatrix}
    L_{\I \I}^{-1}s^{*}_{\I}\\
    L_{\R \I}(L_{\I \I})^{-1}s^{*}_{\I}
    \end{bmatrix}
    =
    \begin{bmatrix}
    L_{\I \I}^{-1}s^{*}_{\I}\\
    K_{\R \I} (K_{\I \I})^{-1}s^{*}_{\I}
    \end{bmatrix}
\end{equation}

\subsubsection{Comparison with the Standard Pivoted Cholesky Decomposition}
\label{asec:pcov_implementation_details_comparison}
In comparison, the standard pivoted Cholesky decomposition starts with the summary statistic ${ s^{(0)}_{\perm{X}} = \diag (K_{\perm{X} \perm{X}}) }$ and iteratively updates it with ${s_{\perm{X}}^{(m+1)} = s_{\perm{X}}^{(m)} - (L_{\perm{X} m})^2}$ which also has $\mathcal{O}(N)$ complexity. 
The overall computational cost of PCov then consists of a single MVP $\mathcal{O}(N^2)$ up front, plus the familiar $\mathcal{O}(M^2N)$ scaling for the iterations of the Cholesky decomposition.

\subsubsection{External Selection}
\label{asec:external_selection}
Another benefit of flexible weighting is that it is easy to include new sources of information in the procedure.
For example, let us assume that we want to learn about a latent function at some inaccessible locations $Z$ by conditioning on possible observations in $X$. The PCov algorithm provides a straightforward implementation to address this question by providing a weighting vector $w = [\zero_{X}, \one_{Z}]\trans$ and considering the combined set of locations $[X,Z]$.
\begin{equation}
    \begin{bmatrix}
      s_X \\
      s_Z
  \end{bmatrix}
  =
  \begin{bmatrix}
  K_{XX} & K_{XZ}\\
  K_{ZX} & K_{ZZ}\\
  \end{bmatrix} 
  \begin{bmatrix}
  \zero_{X} \\ 
  \one_{Z} 
  \end{bmatrix}
  = 
  \begin{bmatrix}
      K_{XZ} \one_Z \\
      K_{ZZ} \one_Z
  \end{bmatrix},
\end{equation}
The point in $X$ that is most informative (under PCov assumptions) is then indicated by the maximal absolute value of $s_X$.

\section{Extended Visual Investigation of the Selection Strategies}
\label{asec:selection_visualization}
\begin{figure}[p]
    \raggedleft

\begin{tikzpicture}[/tikz/background rectangle/.style={fill={rgb,1:red,1.0;green,1.0;blue,1.0}, fill opacity={1.0}, draw opacity={1.0}}, show background rectangle]
\begin{axis}[point meta max={nan}, point meta min={nan}, legend cell align={left}, legend columns={1}, title={}, title style={at={{(0.5,1)}}, anchor={south}, font={{\fontsize{10 pt}{13.0 pt}\selectfont}}, color={rgb,1:red,0.0;green,0.0;blue,0.0}, draw opacity={1.0}, rotate={0.0}, align={center}}, legend style={color={rgb,1:red,0.0;green,0.0;blue,0.0}, draw opacity={0.0}, line width={1}, solid, fill={rgb,1:red,1.0;green,1.0;blue,1.0}, fill opacity={1.0}, text opacity={1.0}, font={{\fontsize{10 pt}{13.0 pt}\selectfont}}, text={rgb,1:red,0.0;green,0.0;blue,0.0}, cells={anchor={center}}, at={(0.98, 0.02)}, anchor={south east}}, axis background/.style={fill={rgb,1:red,1.0;green,1.0;blue,1.0}, opacity={1.0}}, anchor={north west}, xshift={1.0mm}, yshift={-1.0mm}, width={80.55mm}, height={32mm}, scaled x ticks={false}, xlabel={}, x tick style={color={rgb,1:red,0.0;green,0.0;blue,0.0}, opacity={1.0}}, x tick label style={color={rgb,1:red,0.0;green,0.0;blue,0.0}, opacity={1.0}, rotate={0}}, xlabel style={at={(ticklabel cs:0.5)}, anchor=near ticklabel, at={{(ticklabel cs:0.5)}}, anchor={near ticklabel}, font={{\fontsize{10 pt}{13.0 pt}\selectfont}}, color={rgb,1:red,0.0;green,0.0;blue,0.0}, draw opacity={1.0}, rotate={0.0}}, xmajorgrids={false}, xmin={-0.15000000000000036}, xmax={5.15}, xticklabels={{,,,,,}}, xtick={{0.0,1.0,2.0,3.0,4.0,5.0}}, xtick align={inside}, xticklabel style={font={{\fontsize{10 pt}{13.0 pt}\selectfont}}, color={rgb,1:red,0.0;green,0.0;blue,0.0}, draw opacity={1.0}, rotate={0.0}}, x grid style={color={rgb,1:red,0.0;green,0.0;blue,0.0}, draw opacity={0.1}, line width={0.5}, solid}, xticklabel pos={left}, x axis line style={color={rgb,1:red,0.0;green,0.0;blue,0.0}, draw opacity={1.0}, line width={1}, solid}, scaled y ticks={false}, ylabel={y}, y tick style={color={rgb,1:red,0.0;green,0.0;blue,0.0}, opacity={1.0}}, y tick label style={color={rgb,1:red,0.0;green,0.0;blue,0.0}, opacity={1.0}, rotate={0}}, ylabel style={at={(ticklabel cs:0.5)}, anchor=near ticklabel, at={{(ticklabel cs:0.5)}}, anchor={near ticklabel}, font={{\fontsize{10 pt}{13.0 pt}\selectfont}}, color={rgb,1:red,0.0;green,0.0;blue,0.0}, draw opacity={1.0}, rotate={0.0}}, ymajorgrids={false}, ymin={-2.12}, ymax={2.12}, yticklabels={{$-2.0$,$0.0$,$1.8$}}, ytick={{-2.0,0.0,1.8}}, ytick align={inside}, yticklabel style={font={{\fontsize{10 pt}{13.0 pt}\selectfont}}, color={rgb,1:red,0.0;green,0.0;blue,0.0}, draw opacity={1.0}, rotate={0.0}}, y grid style={color={rgb,1:red,0.0;green,0.0;blue,0.0}, draw opacity={0.1}, line width={0.5}, solid}, yticklabel pos={left}, y axis line style={color={rgb,1:red,0.0;green,0.0;blue,0.0}, draw opacity={1.0}, line width={1}, solid}, colorbar={false}]
    \addplot[color={rgb,1:red,1.0;green,0.6471;blue,0.0}, name path={7e4d8fea-56c7-4c5e-992b-2c25add111be}, draw opacity={1.0}, line width={2}, solid, forget plot]
        table[row sep={\\}]
        {
            \\
            1.6  -6.36  \\
            1.6  6.36  \\
        }
        ;
    \addplot[color={rgb,1:red,0.3098;green,0.5804;blue,0.8039}, name path={b25f2207-a66a-4354-8ec2-8325528185e3}, draw opacity={1.0}, line width={2}, dashed, forget plot]
        table[row sep={\\}]
        {
            \\
            3.0  -6.36  \\
            3.0  6.36  \\
        }
        ;
    \addplot[color={rgb,1:red,0.698;green,0.1333;blue,0.1333}, name path={8cdd7072-b8e1-40c8-b004-2d110516c9f3}, draw opacity={1.0}, line width={2}, solid, forget plot]
        table[row sep={\\}]
        {
            \\
            3.0  -6.36  \\
            3.0  6.36  \\
        }
        ;
    \addplot[color={rgb,1:red,0.1804;green,0.5451;blue,0.3412}, name path={80b8b3cf-8e15-4700-adba-749e1adddf29}, draw opacity={1.0}, line width={2}, solid, forget plot]
        table[row sep={\\}]
        {
            \\
            4.75  -6.36  \\
            4.75  6.36  \\
        }
        ;
    \addplot[color={rgb,1:red,0.4902;green,0.149;blue,0.8039}, name path={ab2ce464-9bc2-40a4-970d-65c8f8fe5d22}, draw opacity={1.0}, line width={2}, solid, forget plot]
        table[row sep={\\}]
        {
            \\
            0.35  -6.36  \\
            0.35  6.36  \\
        }
        ;
    \addplot[color={rgb,1:red,0.0;green,0.0;blue,0.0}, name path={bebafb57-f9d1-423e-b764-d92acd408918}, only marks, draw opacity={1.0}, line width={0}, solid, mark={x}, mark size={3.0 pt}, mark repeat={1}, mark options={color={rgb,1:red,0.0;green,0.0;blue,0.0}, draw opacity={1.0}, fill={rgb,1:red,0.0;green,0.0;blue,0.0}, fill opacity={1.0}, line width={0.75}, rotate={0}, solid}]
        table[row sep={\\}]
        {
            \\
            0.1  -0.2  \\
            0.35  -2.0  \\
            0.55  1.8  \\
            1.6  0.0  \\
            2.9  0.0  \\
            3.0  0.0  \\
            3.1  0.0  \\
            4.5  1.0  \\
            4.75  1.0  \\
            5.0  1.0  \\
        }
        ;
    \addlegendentry {data}
    \addplot[color={rgb,1:red,0.0;green,0.0;blue,0.0}, name path={d91fbaa8-26a4-4dbe-8450-0f81039661ab}, only marks, draw opacity={1.0}, line width={0}, solid, mark=*,
    mark size={2.5 pt},
    mark repeat={1}, mark options={color={rgb,1:red,0.0;green,0.0;blue,0.0}, draw opacity={1.0}, fill={rgb,1:red,0.0;green,0.0;blue,0.0}, fill opacity={1.0}, line width={0.75}, rotate={0}, solid}]
        table[row sep={\\}]
        {
            \\
            0.0  0.0  \\
        }
        ;
\end{axis}
\end{tikzpicture}\\[-0.5\baselineskip]

\begin{tikzpicture}[/tikz/background rectangle/.style={fill={rgb,1:red,1.0;green,1.0;blue,1.0}, fill opacity={1.0}, draw opacity={1.0}}, show background rectangle]
\begin{axis}[point meta max={nan}, point meta min={nan}, legend cell align={left}, legend columns={1}, title={}, title style={at={{(0.5,1)}}, anchor={south}, font={{\fontsize{10 pt}{13.0 pt}\selectfont}}, color={rgb,1:red,0.0;green,0.0;blue,0.0}, draw opacity={1.0}, rotate={0.0}, align={center}}, legend style={color={rgb,1:red,0.0;green,0.0;blue,0.0}, draw opacity={0.0}, line width={1}, solid, fill={rgb,1:red,1.0;green,1.0;blue,1.0}, fill opacity={1.0}, text opacity={1.0}, font={{\fontsize{10 pt}{13.0 pt}\selectfont}}, text={rgb,1:red,0.0;green,0.0;blue,0.0}, cells={anchor={center}}, at={(0.98, 0.02)}, anchor={south east}}, axis background/.style={fill={rgb,1:red,1.0;green,1.0;blue,1.0}, opacity={1.0}}, anchor={north west}, xshift={1.0mm}, yshift={-1.0mm}, width={80.55mm}, height={32mm}, scaled x ticks={false}, xlabel={}, x tick style={color={rgb,1:red,0.0;green,0.0;blue,0.0}, opacity={1.0}}, x tick label style={color={rgb,1:red,0.0;green,0.0;blue,0.0}, opacity={1.0}, rotate={0}}, xlabel style={at={(ticklabel cs:0.5)}, anchor=near ticklabel, at={{(ticklabel cs:0.5)}}, anchor={near ticklabel}, font={{\fontsize{10 pt}{13.0 pt}\selectfont}}, color={rgb,1:red,0.0;green,0.0;blue,0.0}, draw opacity={1.0}, rotate={0.0}}, xmajorgrids={false}, xmin={-0.15000000000000036}, xmax={5.15}, xticklabels={{}}, xtick={{0.0,1.0,2.0,3.0,4.0,5.0}}, xtick align={inside}, xticklabel style={font={{\fontsize{10 pt}{13.0 pt}\selectfont}}, color={rgb,1:red,0.0;green,0.0;blue,0.0}, draw opacity={1.0}, rotate={0.0}}, x grid style={color={rgb,1:red,0.0;green,0.0;blue,0.0}, draw opacity={0.1}, line width={0.5}, solid}, xticklabel pos={left}, x axis line style={color={rgb,1:red,0.0;green,0.0;blue,0.0}, draw opacity={1.0}, line width={1}, solid}, scaled y ticks={false}, ylabel={$\alpha(x)$}, y tick style={color={rgb,1:red,0.0;green,0.0;blue,0.0}, opacity={1.0}}, y tick label style={color={rgb,1:red,0.0;green,0.0;blue,0.0}, opacity={1.0}, rotate={0}}, ylabel style={at={(ticklabel cs:0.5)}, anchor=near ticklabel, at={{(ticklabel cs:0.5)}}, anchor={near ticklabel}, font={{\fontsize{10 pt}{13.0 pt}\selectfont}}, color={rgb,1:red,0.0;green,0.0;blue,0.0}, draw opacity={1.0}, rotate={0.0}}, ymajorgrids={false}, ymin={-0.040400000000000234}, ymax={1.0504}, yticklabels={{-0.0, 0.5, 1.0}}, ytick={{-2.220446049250313e-16,0.5049999999999999,1.01}}, ytick align={inside}, yticklabel style={font={{\fontsize{10 pt}{13.0 pt}\selectfont}}, color={rgb,1:red,0.0;green,0.0;blue,0.0}, draw opacity={1.0}, rotate={0.0}}, y grid style={color={rgb,1:red,0.0;green,0.0;blue,0.0}, draw opacity={0.1}, line width={0.5}, solid}, yticklabel pos={left}, y axis line style={color={rgb,1:red,0.0;green,0.0;blue,0.0}, draw opacity={1.0}, line width={1}, solid}, colorbar={false}]
    \addplot[color={rgb,1:red,1.0;green,0.6471;blue,0.0}, name path={b864f59a-7e63-4793-9633-1105b2c837d4}, draw opacity={1.0}, line width={2}, solid, forget plot]
        table[row sep={\\}]
        {
            \\
            1.6  -1.1312000000000004  \\
            1.6  2.1412000000000004  \\
        }
        ;
    \addplot[color={rgb,1:red,1.0;green,0.6471;blue,0.0}, name path={768457e4-55a1-494d-a070-0c3150f188bf}, only marks, draw opacity={1.0}, line width={0}, solid, mark={*}, mark size={3.0 pt}, mark repeat={1}, mark options={color={rgb,1:red,0.0;green,0.0;blue,0.0}, draw opacity={1.0}, fill={rgb,1:red,1.0;green,0.6471;blue,0.0}, fill opacity={1.0}, line width={0.75}, rotate={0}, solid}]
        table[row sep={\\}]
        {
            \\
            0.0  -2.220446049250313e-16  \\
            0.1  0.058723327571957196  \\
            0.35  0.403439213678796  \\
            0.55  0.7147551688812996  \\
            1.6  1.009964640742929  \\
            2.9  1.0099999999999976  \\
            3.0  1.0099999999999998  \\
            3.1  1.01  \\
            4.5  1.01  \\
            4.75  1.01  \\
            5.0  1.01  \\
        }
        ;
    \addlegendentry {Var}
\end{axis}
\end{tikzpicture}\\[-0.5\baselineskip]

\begin{tikzpicture}[/tikz/background rectangle/.style={fill={rgb,1:red,1.0;green,1.0;blue,1.0}, fill opacity={1.0}, draw opacity={1.0}}, show background rectangle]
\begin{axis}[point meta max={nan}, point meta min={nan}, legend cell align={left}, legend columns={1}, title={}, title style={at={{(0.5,1)}}, anchor={south}, font={{\fontsize{10 pt}{13.0 pt}\selectfont}}, color={rgb,1:red,0.0;green,0.0;blue,0.0}, draw opacity={1.0}, rotate={0.0}, align={center}}, legend style={color={rgb,1:red,0.0;green,0.0;blue,0.0}, draw opacity={0.0}, line width={1}, solid, fill={rgb,1:red,1.0;green,1.0;blue,1.0}, fill opacity={1.0}, text opacity={1.0}, font={{\fontsize{10 pt}{13.0 pt}\selectfont}}, text={rgb,1:red,0.0;green,0.0;blue,0.0}, cells={anchor={center}}, at={(0.5, 0.98)}, anchor={north}}, axis background/.style={fill={rgb,1:red,1.0;green,1.0;blue,1.0}, opacity={1.0}}, anchor={north west}, xshift={1.0mm}, yshift={-1.0mm}, width={80.55mm}, height={32mm}, scaled x ticks={false}, xlabel={}, x tick style={color={rgb,1:red,0.0;green,0.0;blue,0.0}, opacity={1.0}}, x tick label style={color={rgb,1:red,0.0;green,0.0;blue,0.0}, opacity={1.0}, rotate={0}}, xlabel style={at={(ticklabel cs:0.5)}, anchor=near ticklabel, at={{(ticklabel cs:0.5)}}, anchor={near ticklabel}, font={{\fontsize{10 pt}{13.0 pt}\selectfont}}, color={rgb,1:red,0.0;green,0.0;blue,0.0}, draw opacity={1.0}, rotate={0.0}}, xmajorgrids={false}, xmin={-0.15000000000000036}, xmax={5.15}, xticklabels={{}}, xtick={{0.0,1.0,2.0,3.0,4.0,5.0}}, xtick align={inside}, xticklabel style={font={{\fontsize{10 pt}{13.0 pt}\selectfont}}, color={rgb,1:red,0.0;green,0.0;blue,0.0}, draw opacity={1.0}, rotate={0.0}}, x grid style={color={rgb,1:red,0.0;green,0.0;blue,0.0}, draw opacity={0.1}, line width={0.5}, solid}, xticklabel pos={left}, x axis line style={color={rgb,1:red,0.0;green,0.0;blue,0.0}, draw opacity={1.0}, line width={1}, solid}, scaled y ticks={false}, ylabel={$\alpha(x)$}, y tick style={color={rgb,1:red,0.0;green,0.0;blue,0.0}, opacity={1.0}}, y tick label style={color={rgb,1:red,0.0;green,0.0;blue,0.0}, opacity={1.0}, rotate={0}}, ylabel style={at={(ticklabel cs:0.5)}, anchor=near ticklabel, at={{(ticklabel cs:0.5)}}, anchor={near ticklabel}, font={{\fontsize{10 pt}{13.0 pt}\selectfont}}, color={rgb,1:red,0.0;green,0.0;blue,0.0}, draw opacity={1.0}, rotate={0.0}}, ymajorgrids={false}, ymin={-0.08}, ymax={2.08}, yticklabels={{ 0.0, 1.0, 2.0}}, ytick={{0.0,1.0,2.0}}, ytick align={inside}, yticklabel style={font={{\fontsize{10 pt}{13.0 pt}\selectfont}}, color={rgb,1:red,0.0;green,0.0;blue,0.0}, draw opacity={1.0}, rotate={0.0}}, y grid style={color={rgb,1:red,0.0;green,0.0;blue,0.0}, draw opacity={0.1}, line width={0.5}, solid}, yticklabel pos={left}, y axis line style={color={rgb,1:red,0.0;green,0.0;blue,0.0}, draw opacity={1.0}, line width={1}, solid}, colorbar={false}]
    \addplot[color={rgb,1:red,0.4902;green,0.149;blue,0.8039}, name path={7738a350-2664-4a42-9006-bcec66da76ff}, draw opacity={1.0}, line width={2}, solid, forget plot]
        table[row sep={\\}]
        {
            \\
            0.35  -2.24  \\
            0.35  4.24  \\
        }
        ;
    \addplot[color={rgb,1:red,0.4902;green,0.149;blue,0.8039}, name path={3554a758-c55f-4c01-8fc2-d780c142ca8c}, only marks, draw opacity={1.0}, line width={0}, solid, mark={square*}, mark size={3.0 pt}, mark repeat={1}, mark options={color={rgb,1:red,0.0;green,0.0;blue,0.0}, draw opacity={1.0}, fill={rgb,1:red,0.4902;green,0.149;blue,0.8039}, fill opacity={1.0}, line width={0.75}, rotate={0}, solid}]
        table[row sep={\\}]
        {
            \\
            0.0  0.0  \\
            0.1  0.2  \\
            0.35  2.0  \\
            0.55  1.8  \\
            1.6  0.0  \\
            2.9  0.0  \\
            3.0  0.0  \\
            3.1  0.0  \\
            4.5  1.0  \\
            4.75  1.0  \\
            5.0  1.0  \\
        }
        ;
    \addlegendentry {ME}
\end{axis}
\end{tikzpicture}\hspace{-3pt}\\[-0.5\baselineskip]

\begin{tikzpicture}[/tikz/background rectangle/.style={fill={rgb,1:red,1.0;green,1.0;blue,1.0}, fill opacity={1.0}, draw opacity={1.0}}, show background rectangle]
\begin{axis}[point meta max={nan}, point meta min={nan}, legend cell align={left}, legend columns={1}, title={}, title style={at={{(0.5,1)}}, anchor={south}, font={{\fontsize{10 pt}{13.0 pt}\selectfont}}, color={rgb,1:red,0.0;green,0.0;blue,0.0}, draw opacity={1.0}, rotate={0.0}, align={center}}, legend style={color={rgb,1:red,0.0;green,0.0;blue,0.0}, draw opacity={0.0}, line width={1}, solid, fill={rgb,1:red,1.0;green,1.0;blue,1.0}, fill opacity={1.0}, text opacity={1.0}, font={{\fontsize{10 pt}{13.0 pt}\selectfont}}, text={rgb,1:red,0.0;green,0.0;blue,0.0}, cells={anchor={center}}, at={(0.02, 0.98)}, anchor={north west}}, axis background/.style={fill={rgb,1:red,1.0;green,1.0;blue,1.0}, opacity={1.0}}, anchor={north west}, xshift={1.0mm}, yshift={-1.0mm}, width={80.55mm}, height={32mm}, scaled x ticks={false}, xlabel={}, x tick style={color={rgb,1:red,0.0;green,0.0;blue,0.0}, opacity={1.0}}, x tick label style={color={rgb,1:red,0.0;green,0.0;blue,0.0}, opacity={1.0}, rotate={0}}, xlabel style={at={(ticklabel cs:0.5)}, anchor=near ticklabel, at={{(ticklabel cs:0.5)}}, anchor={near ticklabel}, font={{\fontsize{10 pt}{13.0 pt}\selectfont}}, color={rgb,1:red,0.0;green,0.0;blue,0.0}, draw opacity={1.0}, rotate={0.0}}, xmajorgrids={false}, xmin={-0.15000000000000036}, xmax={5.15}, xticklabels={{}}, xtick={{0.0,1.0,2.0,3.0,4.0,5.0}}, xtick align={inside}, xticklabel style={font={{\fontsize{10 pt}{13.0 pt}\selectfont}}, color={rgb,1:red,0.0;green,0.0;blue,0.0}, draw opacity={1.0}, rotate={0.0}}, x grid style={color={rgb,1:red,0.0;green,0.0;blue,0.0}, draw opacity={0.1}, line width={0.5}, solid}, xticklabel pos={left}, x axis line style={color={rgb,1:red,0.0;green,0.0;blue,0.0}, draw opacity={1.0}, line width={1}, solid}, scaled y ticks={false}, ylabel={$\alpha(x)$}, y tick style={color={rgb,1:red,0.0;green,0.0;blue,0.0}, opacity={1.0}}, y tick label style={color={rgb,1:red,0.0;green,0.0;blue,0.0}, opacity={1.0}, rotate={0}}, ylabel style={at={(ticklabel cs:0.5)}, anchor=near ticklabel, at={{(ticklabel cs:0.5)}}, anchor={near ticklabel}, font={{\fontsize{10 pt}{13.0 pt}\selectfont}}, color={rgb,1:red,0.0;green,0.0;blue,0.0}, draw opacity={1.0}, rotate={0.0}}, ymajorgrids={false}, ymin={-1.4936742040519027}, ymax={65.83552930534924}, yticklabels={{ 1.0, 32.2, 63.3}}, ytick={{0.9999999999999918,32.17092755064867,63.34185510129735}}, ytick align={inside}, yticklabel style={font={{\fontsize{10 pt}{13.0 pt}\selectfont}}, color={rgb,1:red,0.0;green,0.0;blue,0.0}, draw opacity={1.0}, rotate={0.0}}, y grid style={color={rgb,1:red,0.0;green,0.0;blue,0.0}, draw opacity={0.1}, line width={0.5}, solid}, yticklabel pos={left}, y axis line style={color={rgb,1:red,0.0;green,0.0;blue,0.0}, draw opacity={1.0}, line width={1}, solid}, colorbar={false}]
    \addplot[color={rgb,1:red,0.698;green,0.1333;blue,0.1333}, name path={3e45f546-e639-491d-b2b8-773b3dc34045}, draw opacity={1.0}, line width={2}, solid, forget plot]
        table[row sep={\\}]
        {
            \\
            3.0  -68.82287771345304  \\
            3.0  133.16473281475038  \\
        }
        ;
    \addplot[color={rgb,1:red,0.698;green,0.1333;blue,0.1333}, name path={a0900774-1175-45c4-9bc7-b5cc761ad4ee}, only marks, draw opacity={1.0}, line width={0}, solid, mark={pentagon*}, mark size={3.0 pt}, mark repeat={1}, mark options={color={rgb,1:red,0.0;green,0.0;blue,0.0}, draw opacity={1.0}, fill={rgb,1:red,0.698;green,0.1333;blue,0.1333}, fill opacity={1.0}, line width={0.75}, rotate={0}, solid}]
        table[row sep={\\}]
        {
            \\
            0.0  0.9999999999999918  \\
            0.1  3.117192774710076  \\
            0.35  13.148525770640898  \\
            0.55  8.776474873429917  \\
            1.6  1.0456082707638619  \\
            2.9  22.396959401615497  \\
            3.0  63.34185510129735  \\
            3.1  22.42004007683284  \\
            4.5  8.089123079936671  \\
            4.75  21.82110857947968  \\
            5.0  8.058085494169108  \\
        }
        ;
    \addlegendentry {MI}
\end{axis}
\end{tikzpicture}\\[-0.5\baselineskip]

\begin{tikzpicture}[/tikz/background rectangle/.style={fill={rgb,1:red,1.0;green,1.0;blue,1.0}, fill opacity={1.0}, draw opacity={1.0}}, show background rectangle]
\begin{axis}[point meta max={nan}, point meta min={nan}, legend cell align={left}, legend columns={1}, title={}, title style={at={{(0.5,1)}}, anchor={south}, font={{\fontsize{10 pt}{13.0 pt}\selectfont}}, color={rgb,1:red,0.0;green,0.0;blue,0.0}, draw opacity={1.0}, rotate={0.0}, align={center}}, legend style={color={rgb,1:red,0.0;green,0.0;blue,0.0}, draw opacity={0.0}, line width={1}, solid, fill={rgb,1:red,1.0;green,1.0;blue,1.0}, fill opacity={1.0}, text opacity={1.0}, font={{\fontsize{10 pt}{13.0 pt}\selectfont}}, text={rgb,1:red,0.0;green,0.0;blue,0.0}, cells={anchor={center}}, at={(0.02, 0.98)}, anchor={north west}}, axis background/.style={fill={rgb,1:red,1.0;green,1.0;blue,1.0}, opacity={1.0}}, anchor={north west}, xshift={1.0mm}, yshift={-1.0mm}, width={80.55mm}, height={32mm}, scaled x ticks={false}, xlabel={}, x tick style={color={rgb,1:red,0.0;green,0.0;blue,0.0}, opacity={1.0}}, x tick label style={color={rgb,1:red,0.0;green,0.0;blue,0.0}, opacity={1.0}, rotate={0}}, xlabel style={at={(ticklabel cs:0.5)}, anchor=near ticklabel, at={{(ticklabel cs:0.5)}}, anchor={near ticklabel}, font={{\fontsize{10 pt}{13.0 pt}\selectfont}}, color={rgb,1:red,0.0;green,0.0;blue,0.0}, draw opacity={1.0}, rotate={0.0}}, xmajorgrids={false}, xmin={-0.15000000000000036}, xmax={5.15}, xticklabels={{}}, xtick={{0.0,1.0,2.0,3.0,4.0,5.0}}, xtick align={inside}, xticklabel style={font={{\fontsize{10 pt}{13.0 pt}\selectfont}}, color={rgb,1:red,0.0;green,0.0;blue,0.0}, draw opacity={1.0}, rotate={0.0}}, x grid style={color={rgb,1:red,0.0;green,0.0;blue,0.0}, draw opacity={0.1}, line width={0.5}, solid}, xticklabel pos={left}, x axis line style={color={rgb,1:red,0.0;green,0.0;blue,0.0}, draw opacity={1.0}, line width={1}, solid}, scaled y ticks={false}, ylabel={$\alpha(x)$}, y tick style={color={rgb,1:red,0.0;green,0.0;blue,0.0}, opacity={1.0}}, y tick label style={color={rgb,1:red,0.0;green,0.0;blue,0.0}, opacity={1.0}, rotate={0}}, ylabel style={at={(ticklabel cs:0.5)}, anchor=near ticklabel, at={{(ticklabel cs:0.5)}}, anchor={near ticklabel}, font={{\fontsize{10 pt}{13.0 pt}\selectfont}}, color={rgb,1:red,0.0;green,0.0;blue,0.0}, draw opacity={1.0}, rotate={0.0}}, ymajorgrids={false}, ymin={-0.12015509259223944}, ymax={3.1240324073982073}, yticklabels={{-0.0, 1.5, 3.0}}, ytick={{-6.678685490044187e-16,1.501938657402984,3.0038773148059685}}, ytick align={inside}, yticklabel style={font={{\fontsize{10 pt}{13.0 pt}\selectfont}}, color={rgb,1:red,0.0;green,0.0;blue,0.0}, draw opacity={1.0}, rotate={0.0}}, y grid style={color={rgb,1:red,0.0;green,0.0;blue,0.0}, draw opacity={0.1}, line width={0.5}, solid}, yticklabel pos={left}, y axis line style={color={rgb,1:red,0.0;green,0.0;blue,0.0}, draw opacity={1.0}, line width={1}, solid}, colorbar={false}]
    \addplot[color={rgb,1:red,0.3098;green,0.5804;blue,0.8039}, name path={f6be8641-ef21-4e45-937f-3721c50637a0}, draw opacity={1.0}, line width={2}, solid, forget plot]
        table[row sep={\\}]
        {
            \\
            3.0  -3.3643425925826858  \\
            3.0  6.368219907388654  \\
        }
        ;
    \addplot[color={rgb,1:red,0.3098;green,0.5804;blue,0.8039}, name path={578e2a64-794f-4aa1-9431-1f6531045a63}, only marks, draw opacity={1.0}, line width={0}, solid, mark={triangle*}, mark size={3.0 pt}, mark repeat={1}, mark options={color={rgb,1:red,0.0;green,0.0;blue,0.0}, draw opacity={1.0}, fill={rgb,1:red,0.3098;green,0.5804;blue,0.8039}, fill opacity={1.0}, line width={0.75}, rotate={180}, solid}]
        table[row sep={\\}]
        {
            \\
            0.0  -6.678685490044187e-16  \\
            0.1  0.32393483541516765  \\
            0.35  1.0655685553760823  \\
            0.55  1.458747189174003  \\
            1.6  1.2265968094920745  \\
            2.9  2.9545692691326013  \\
            3.0  3.0038773148059685  \\
            3.1  2.9493172773405303  \\
            4.5  2.5359537260389295  \\
            4.75  2.7825639049383977  \\
            5.0  2.5002425757952587  \\
        }
        ;
    \addlegendentry {PCov}
\end{axis}
\end{tikzpicture}\\[-0.5\baselineskip]

\begin{tikzpicture}[/tikz/background rectangle/.style={fill={rgb,1:red,1.0;green,1.0;blue,1.0}, fill opacity={1.0}, draw opacity={1.0}}, show background rectangle]
\begin{axis}[point meta max={nan}, point meta min={nan}, legend cell align={left}, legend columns={1}, title={}, title style={at={{(0.5,1)}}, anchor={south}, font={{\fontsize{10 pt}{13.0 pt}\selectfont}}, color={rgb,1:red,0.0;green,0.0;blue,0.0}, draw opacity={1.0}, rotate={0.0}, align={center}}, legend style={color={rgb,1:red,0.0;green,0.0;blue,0.0}, draw opacity={0.0}, line width={1}, solid, fill={rgb,1:red,1.0;green,1.0;blue,1.0}, fill opacity={1.0}, text opacity={1.0}, font={{\fontsize{10 pt}{13.0 pt}\selectfont}}, text={rgb,1:red,0.0;green,0.0;blue,0.0}, cells={anchor={center}}, at={(0.02, 0.98)}, anchor={north west}}, axis background/.style={fill={rgb,1:red,1.0;green,1.0;blue,1.0}, opacity={1.0}}, anchor={north west}, xshift={1.0mm}, yshift={-1.0mm}, width={80.55mm}, height={32mm}, scaled x ticks={false}, xlabel={$x$}, x tick style={color={rgb,1:red,0.0;green,0.0;blue,0.0}, opacity={1.0}}, x tick label style={color={rgb,1:red,0.0;green,0.0;blue,0.0}, opacity={1.0}, rotate={0}}, xlabel style={at={(ticklabel cs:0.5)}, anchor=near ticklabel, at={{(ticklabel cs:0.5)}}, anchor={near ticklabel}, font={{\fontsize{10 pt}{13.0 pt}\selectfont}}, color={rgb,1:red,0.0;green,0.0;blue,0.0}, draw opacity={1.0}, rotate={0.0}}, xmajorgrids={false}, xmin={-0.15000000000000036}, xmax={5.15}, xticklabels={{$0$,$1$,$2$,$3$,$4$,$5$}}, xtick={{0.0,1.0,2.0,3.0,4.0,5.0}}, xtick align={inside}, xticklabel style={font={{\fontsize{10 pt}{13.0 pt}\selectfont}}, color={rgb,1:red,0.0;green,0.0;blue,0.0}, draw opacity={1.0}, rotate={0.0}}, x grid style={color={rgb,1:red,0.0;green,0.0;blue,0.0}, draw opacity={0.1}, line width={0.5}, solid}, xticklabel pos={left}, x axis line style={color={rgb,1:red,0.0;green,0.0;blue,0.0}, draw opacity={1.0}, line width={1}, solid}, scaled y ticks={false}, ylabel={$\alpha(x)$}, y tick style={color={rgb,1:red,0.0;green,0.0;blue,0.0}, opacity={1.0}}, y tick label style={color={rgb,1:red,0.0;green,0.0;blue,0.0}, opacity={1.0}, rotate={0}}, ylabel style={at={(ticklabel cs:0.5)}, anchor=near ticklabel, at={{(ticklabel cs:0.5)}}, anchor={near ticklabel}, font={{\fontsize{10 pt}{13.0 pt}\selectfont}}, color={rgb,1:red,0.0;green,0.0;blue,0.0}, draw opacity={1.0}, rotate={0.0}}, ymajorgrids={false}, ymin={-0.11099975220676761}, ymax={2.8859935573759596}, yticklabels={{ 0.0, 1.4, 2.8}}, ytick={{6.66133814775094e-17,1.387496902584596,2.774993805169192}}, ytick align={inside}, yticklabel style={font={{\fontsize{10 pt}{13.0 pt}\selectfont}}, color={rgb,1:red,0.0;green,0.0;blue,0.0}, draw opacity={1.0}, rotate={0.0}}, y grid style={color={rgb,1:red,0.0;green,0.0;blue,0.0}, draw opacity={0.1}, line width={0.5}, solid}, yticklabel pos={left}, y axis line style={color={rgb,1:red,0.0;green,0.0;blue,0.0}, draw opacity={1.0}, line width={1}, solid}, colorbar={false}]
    \addplot[color={rgb,1:red,0.1804;green,0.5451;blue,0.3412}, name path={a415d354-3945-4cad-8908-3c1483e4a10d}, draw opacity={1.0}, line width={2}, solid, forget plot]
        table[row sep={\\}]
        {
            \\
            4.75  -3.107993061789495  \\
            4.75  5.882986866958687  \\
        }
        ;
    \addplot[color={rgb,1:red,0.1804;green,0.5451;blue,0.3412}, name path={ab521afd-fa5c-45f4-9246-ab2579283d08}, only marks, draw opacity={1.0}, line width={0}, solid, mark={diamond*}, mark size={3.0 pt}, mark repeat={1}, mark options={color={rgb,1:red,0.0;green,0.0;blue,0.0}, draw opacity={1.0}, fill={rgb,1:red,0.1804;green,0.5451;blue,0.3412}, fill opacity={1.0}, line width={0.75}, rotate={0}, solid}]
        table[row sep={\\}]
        {
            \\
            0.0  6.66133814775094e-17  \\
            0.1  0.010891766962557158  \\
            0.35  0.06842400422288297  \\
            0.55  0.25928981300303655  \\
            1.6  0.11296168999305085  \\
            2.9  0.00721278680023316  \\
            3.0  0.013641366654505039  \\
            3.1  0.024894245975744612  \\
            4.5  2.499027562297277  \\
            4.75  2.774993805169192  \\
            5.0  2.499027562297229  \\
        }
        ;
    \addlegendentry {WPCov}
\end{axis}
\end{tikzpicture}\hspace{-3pt}
	\caption{%
        A visualization of the different selection strategies on a contrived 1-dimensional data set, as explained in Section~\ref{sec:theory}.
        All strategies share the leftmost point (black circle) as an initial observation as well as the same exponentiated quadratic kernel and a zero mean prior.
        The vertical bars highlight which data point is sampled next for pivoting during the pivoted Cholesky decomposition.
        The methods either chose the next point based on: the posterior variance (orange circles), the maximum squared error of the prediction (purple squares), the mutual information (red pentagons), a combination of the posterior variance and the mutual information (blue triangles), or the previous quantity weighted with the initial residual (green diamonds).
        The last two selection strategies are introduced in this paper.}
	\label{afig:selection_strategies_all}
\end{figure}
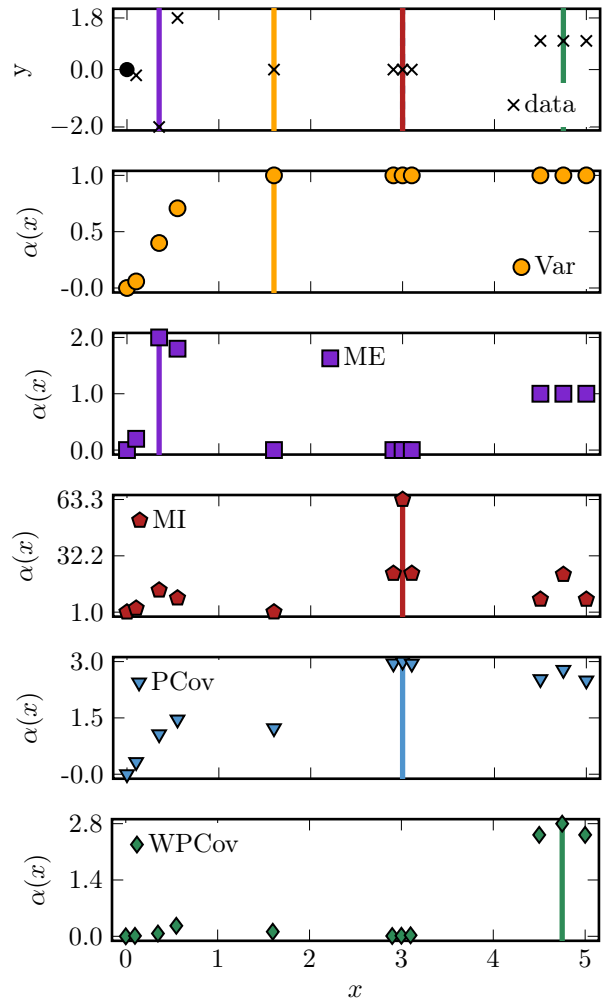

The example in Figure~\ref{afig:selection_strategies_all}, which is the extension of Figure~\ref{fig:selection_strategies_overview}, has been hand-crafted in order to highlight the differences among the selection strategies discussed in the paper. 
We can see how each algorithm would behave when presented with the visible data at the top and where the first point to the left has already been observed ($f(x_1)$). The data is ordered from left to right and the kernel used for this setting is a squared exponential kernel with length scale $\ell =0.5$.

The standard pivoted Cholesky decomposition, here denoted as Var by reasons explained in \ref{asec:d-optimal}, gets a higher acquisition function the further away from $x_1$ we move. However, it stagnates to a value within machine precision beyond the selected point $x_5$, and any point beyond that would be a valid selection. Because the data is ordered from left to right the first occurrence which is indistinguishable from the true optimum is selected.

The Residual selection simply picks points that have a large residual error. WPCov on the other hand weights each residual with the accompanying feature (a Gaussian distribution centered on the point), so it also takes into account how well the residual matches the model.
In this case, it means that WPCov favors the second to last point $x_{10}$ over Residual's $x_3$ because there are several points with a positive residual nearby so there is a better chance of reducing the MSE there.

Both MI and PCov select the same point because it is close to additional points and we do not have information in that area yet. The first 4 points are also clustered together and would be good candidates but by observing $x_1$ we have significantly reduced the information among those points under the presented model.

\clearpage
\section{Trace Reduction of PCov}
\label{asec:trace_estimation}
In this section, we will explore why the PCov selection is a good proxy for a trace-minimizing selection criterion. We will start by looking at the governing equations for trace reduction (Section~\ref{asec:trace_general}) and then specialize the selection to the target setting of covariance matrices arising from stationary kernels, culminating in a lower (Section~\ref{asec:trace_lower}) as well as upper bound (Section~\ref{asec:trace_upper}).
An empirical analysis of the bounds is available in Section~\ref{asec:trace_empirical}.

\subsection{Trace of a Symmetric Matrix}
\label{asec:trace_general}
Consider a generic symmetric and positive definite matrix $G\in \Re^{N\times N}$. We will reuse the notation of the main paper and let $X$ be shorthand for a complete column/row of the matrix so $G_{iX}$ denotes the $i$-th row.
The trace is defined as
\begin{equation}
    \tr (G) = \sum\limits_{j=1}^N G_{jj}
\end{equation}
and the reduction in trace resulting from selecting pivot $i$ is 
\begin{equation}
	\tau_i = \tr(L_{Xi}L_{Xi}\trans)
    = \frac{ \norm{ G_{iX} }^2}{G_{ii}}
    = \frac{1}{G_{ii}}\sum\limits_{j=1}^{N}G_{ij}^2
    \label{aeq:trace_reduction}
\end{equation}
The main trick to deriving our bounds is to consider the trace reduction in~\eqref{aeq:trace_reduction} as the second moment of a random variable, which we can decompose in a mean and variance.
To this end, we introduce the following quantities 
\begin{align}
    s_i &= \frac{1}{N}\sum\limits_{j=1}^{N} G_{ij}^2, \\
    m_i &= \frac{1}{N}\sum\limits_{j=1}^{N} G_{ij}, \\
    v_i &= \frac{1}{N}\sum\limits_{j=1}^{N} (G_{ij} - m_i)^2 = \frac{1}{N}\sum\limits_{j=1}^{N} G^2_{ij} - m_i^2.
\end{align}
Next, we rescale the trace reduction~\eqref{aeq:trace_reduction} by a factor $1/N$ (since all factors are scaled equal it does not change which element is the largest) and rewrite it in terms of the new quantities.
\begin{align}
	\tau_i / N
    &=\frac{1}{N}\frac{1}{G_{ii}}\sum\limits_{j=1}^{N}G_{ij}^2 
        = \frac{1}{G_{ii}} ( s_i ) \notag\\
    &= \frac{1}{G_{ii}} ( m_i^2 + \underbrace{s_i- m_i^2}_{v_i} ) 
        = \frac{1}{G_{ii}} \left( m_i^2 + v_i \right) \label{aeq:trace_moment_decomposition}
\end{align}
This result shows that the trace reduction can be expressed as the sum of the squared mean and variance of each row. 
Since the mean minimizes the squared distance to all points it will minimize the variance term $v_i$. 
From this observation, we can draw the following two conclusions. First, we can use $m_i^2$ as a cheap proxy to estimate how much a pivot would reduce the trace. Second, selecting a pivot based on the maximum $m_i$ will be a good proxy if $v_i$ is small, i.e., the variance is low compared to the mean. 

\textbf{Remark.} One can express $s_i$ in terms of an arbitrary shift $c_i$
\begin{align}
	N s_i
    &= \sum\limits_j \left(G_{ij} \right)^2 \notag\\
    &= \left(-N c_i^2  + 2c_i N m_i\right) + \sum\limits_j \left(G_{ij}- c_i \right)^2_j,
\end{align}
which could be useful for different kinds of estimators but was not further explored. 



\begin{figure*}[t]
    \centering
    \includegraphics[width=0.31\linewidth]{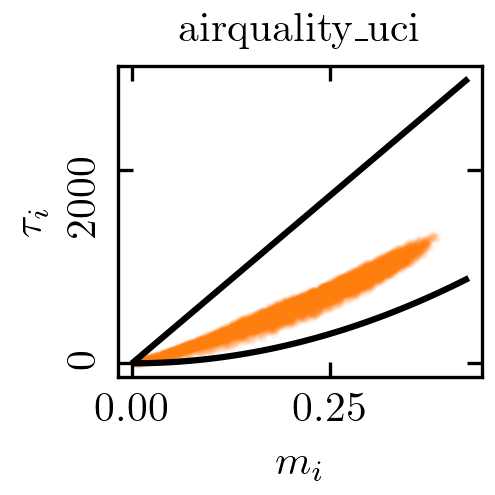}
    \hfill
    \includegraphics[width=0.31\linewidth]{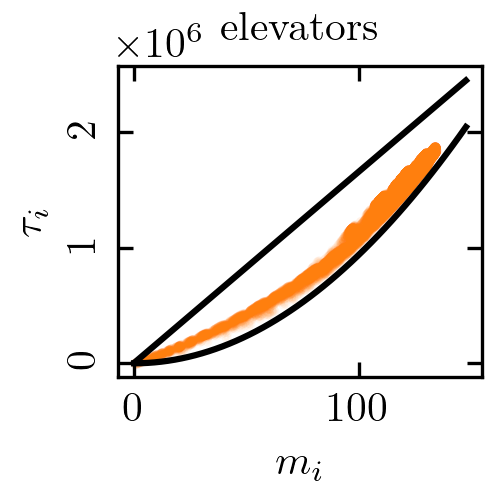}
    \hfill
    \includegraphics[width=0.31\linewidth]{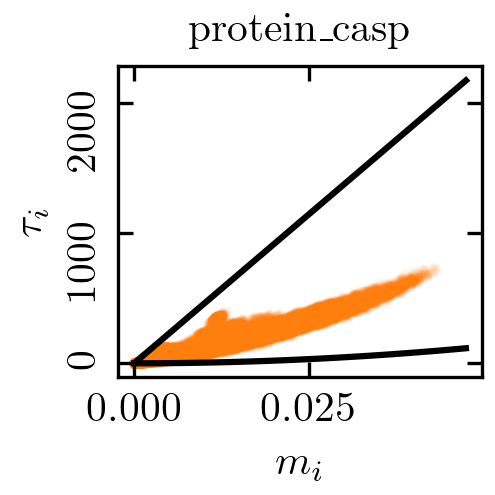}
    \caption{Scatter plot for $m_i$ vs trace reduction with the bounds from Sections~\ref{asec:trace_lower}~and~\ref{asec:trace_upper} overlaid.
    There is a clear correlation between $m_i$ and $tr(i)$, which indicates that a selection based on $m_i$ will also result in a large trace reduction.}
    \label{afig:trace_bounds}
\end{figure*}

\subsection{Trace - Lower Bound}
\label{asec:trace_lower}
A generic lower bound on the trace reduction is readily available in~\eqref{aeq:trace_moment_decomposition}.
Since the variance is always positive we find a lower bound on the trace reduction for $v_i=0$.
\begin{align}
    \tau_i/N &\geq \frac{1}{G_{ii}} m_i^2 \text{ or equivalent} \notag\\
    \tau_i &\geq \frac{m_i}{G_{ii}}\sum\limits_{j=1}^N G_{ij} 
    \label{aeq:trace_lower_bounds}
\end{align}

\subsection{Trace - Upper bound}
\label{asec:trace_upper}
Upper-bounding the variance requires additional information about the matrix.
For the special case of Gram matrices arising from stationary kernels, there are a few additional properties that can be used to further bound the trace reduction.
First, all entries are assumed positive which holds for most stationary kernels.
Second, we know that the maximum value of each row occurs at the diagonal (we additionally know that the diagonal elements are identical).
In other words, this means $0\leq G_{ij} \leq G_{ii}$ for any $j\in {1,...,N}$.

We can then upper bound the variance by considering the worst-case scenario. 
For a row with mean $m_i$ the maximal possible variance then occurs if a fraction $m_i/G_{ii}$ of the values in $G_{iX}$ have the value $G_{ii}$ and the remaining points have value $0$. The resulting upper bound on the variance is
\begin{align}
    v_i 
    &\leq  \frac{m_i}{G_{ii}}(G_{ii} - m_i)^2 + \left(1 - \frac{m_i}{G_{ii}} \right) m_i^2 \notag\\
    &=  \frac{m_i}{G_{ii}}(G^2_{ii} - 2G_{ii}m_i  + m_i^2 + m_i^2 - \frac{m_i}{G_{ii}}m_i^2) \notag\\
    &=  (m_i G_{ii} - 2m_i^2  + \frac{m_i}{G_{ii}}m_i^2 + m_i^2 - \frac{m_i}{G_{ii}}m_i^2) \notag\\
    &=  (m_i G_{ii} - m_i^2 ) = m_i(G_{ii}-m_i) \label{aeq:variance_upper_bound}.
\end{align}
Replacing $v_i$ in~\eqref{aeq:trace_moment_decomposition} with~\eqref{aeq:variance_upper_bound} results in the following upper bound on the trace reduction associated with row $i$
\begin{equation}
     \tau_i/N \leq \frac{1}{G_{ii}}\left( m_i^2 + m_i(G_{ii} - m_i) \right) = m_i\\
\end{equation}
or equivalently
\begin{equation}
     \tau_i \leq \frac{N}{G_{ii}}\left( m_i^2 + m_i(G_{ii} - m_i) \right) = N m_i = \sum\limits_{j=1}^{N}G_{ij},
    \label{aeq:trace_upper_bounds}
\end{equation}
where the last term is the selection criterion of PCov (Section~\ref{sec:pcov}).




\subsection{Empirical Evaluation}
\label{asec:trace_empirical}
Combining the bounds from~\eqref{aeq:trace_lower_bounds} and~\eqref{aeq:trace_upper_bounds} we obtain the following statement about the trace reduction
\begin{equation}
    \frac{m_i^2}{G_{ii}} \leq \tau_i/N \leq  m_i.
    \label{aeq:trace_bounds}
\end{equation}
When the diagonal elements are all identical we see that bounds are only influenced by $m_i$ so large values will result in higher upper as well as lower bounds for the trace reduction.

The PCov selection rule is not optimal in terms of trace reduction but we can empirically assess the correlation between $\tau_i/N$ and $m_i$.
In Figure\ref{afig:trace_bounds} we present a scatter plot for these two quantities for three different data sets along with the bounds of~\eqref{aeq:trace_bounds}.

\section{Further Details on the Preconditioned Conjugate Gradients Experiments}
\label{asec:cg_convergence}
As described in Section~\ref{sec:cg_convergence}, for every data set, which has been obtained from~\cite{Dua_ML_datasets}, the linear system of equations $G \alpha = y$ is solved for $\alpha$ using preconditioned conjugate gradients.
We did not improve the condition of $G$ to ensure that the experiment's results are obfuscated as least as possible.

\begin{figure*}[t] 
    \centering
    \hspace*{3.0em}\includegraphics{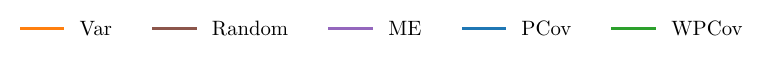}\\[-1em]
    \includegraphics{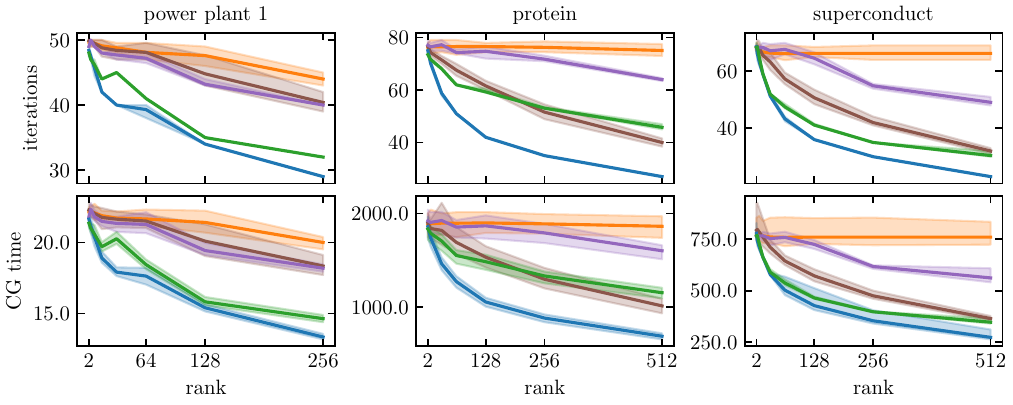}\\
    \includegraphics{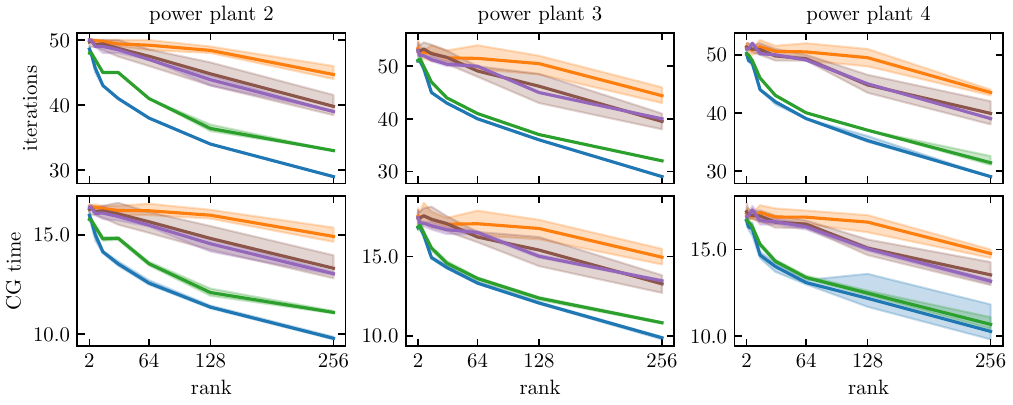}
    \caption{%
        Results of the preconditioned conjugate gradient experiments on all twelve data sets (columns).
        The performance of all methods is measured by the number of iterations until convergence, as well as wall clock runtime including data selection for preconditioner  (rows).
        Lower values are better.
        The shaded areas mark the 90\% confidence interval originating from 10 repetitions with random permutations of the data set for all methods.
        }
	\label{afig:cg_convergence_results_all}
\end{figure*}
\begin{figure*}[t]
    \ContinuedFloat
    \centering
    \hspace*{3.0em}\includegraphics{fig/cg_convergence/legend.pdf}\\[-1em]
    \includegraphics{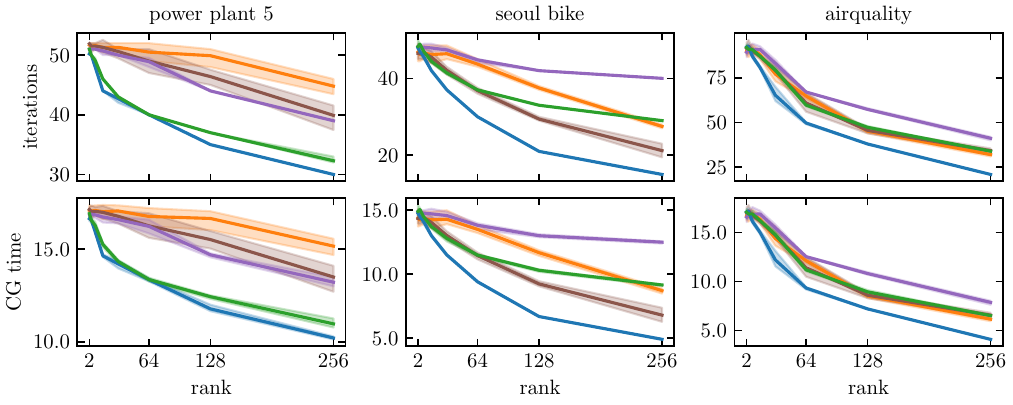}\\
    \includegraphics{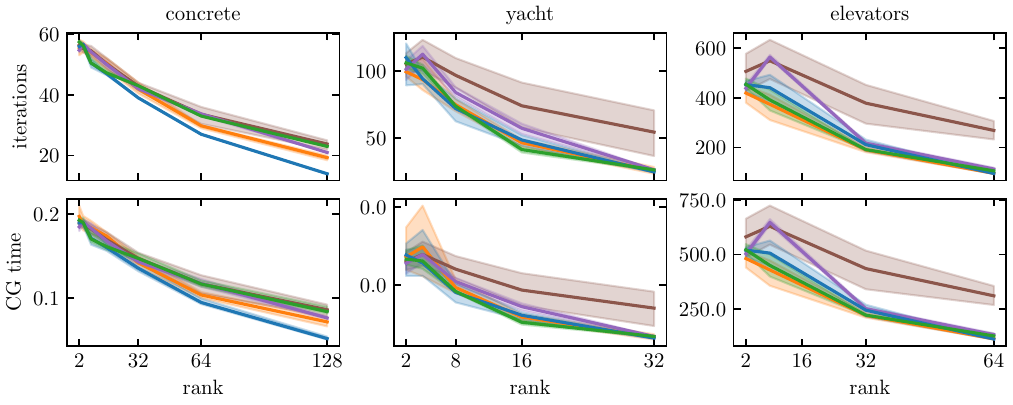}
	\caption{(continued)}
\end{figure*}

\section{Further Details on the Sparse Gaussian Process Regression Experiments}
\label{asec:gp_regression}
As described in Section~\ref{sec:gp_regression}, one Gaussian process is trained for every data set, which has been obtained from~\cite{Dua_ML_datasets}.
For training the GPs, we employed the Julia packages KernelFunctions.jl, AbstractGPs.jl, Zygote.jl, and Optim.jl, using the LBFGS optimizer and a squared exponential kernel for all methods.
The data sets \dsname{yacht}, \dsname{concrete}, and \dsname{airquality} have 308, 1030, and 6941, samples, i.e., less than 7000, thus, are not affected by the truncation described in Section~\ref{sec:experiments}.

\begin{figure*}[t] 
    \centering
    \hspace*{3.0em}\includegraphics{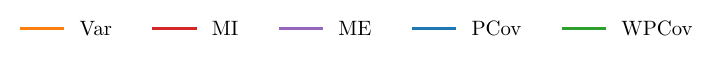}\\[-1em]
    \includegraphics{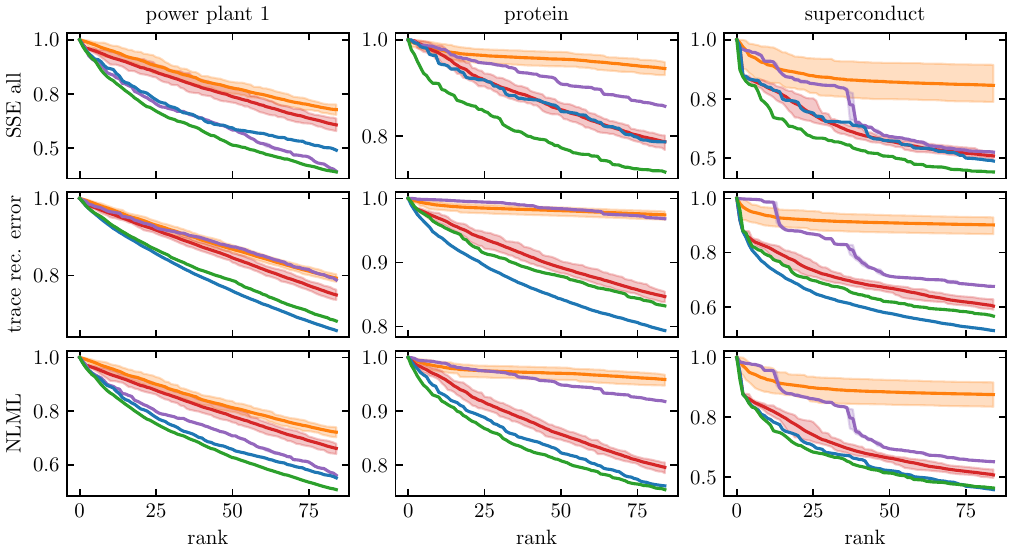}\\
    \includegraphics{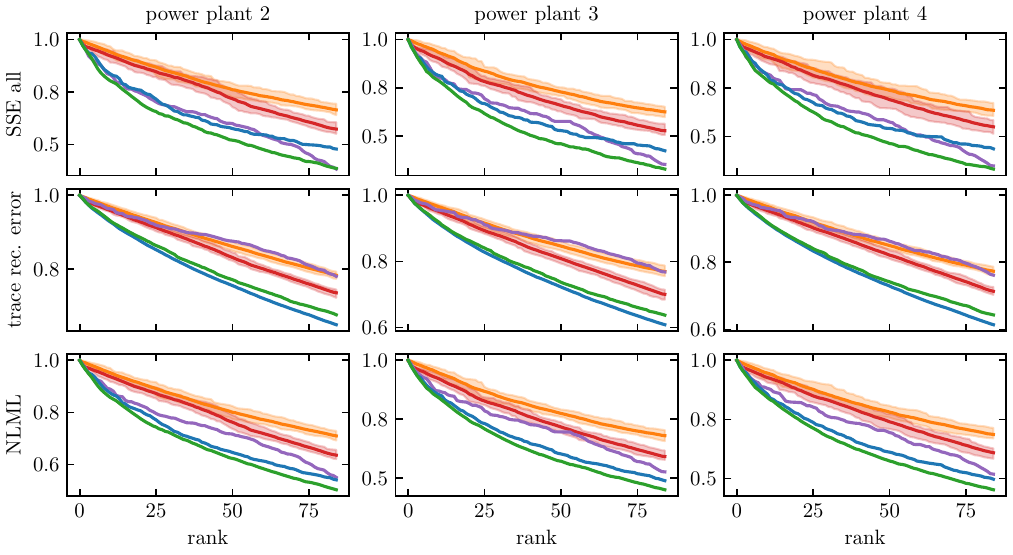}
    \caption{%
        Results of the sparse GP regression experiments on all twelve data sets (columns).
        The performance of all methods is measured on the metrics (rows) described in Section~\ref{sec:gp_regression}.
        Lower values are better for all metrics, which were normalized to facilitate comparison among the data sets.
        The shaded areas mark the 90\% confidence interval originating from 20 repetitions with random permutations of the data set for all methods.
        }
	\label{afig:gp_regression_results_all}
\end{figure*}
\begin{figure*}[t] 
    \ContinuedFloat
    \centering
    \hspace*{3.0em}\includegraphics{fig/gp_regression/legend.pdf}\\[-1em]
    \includegraphics{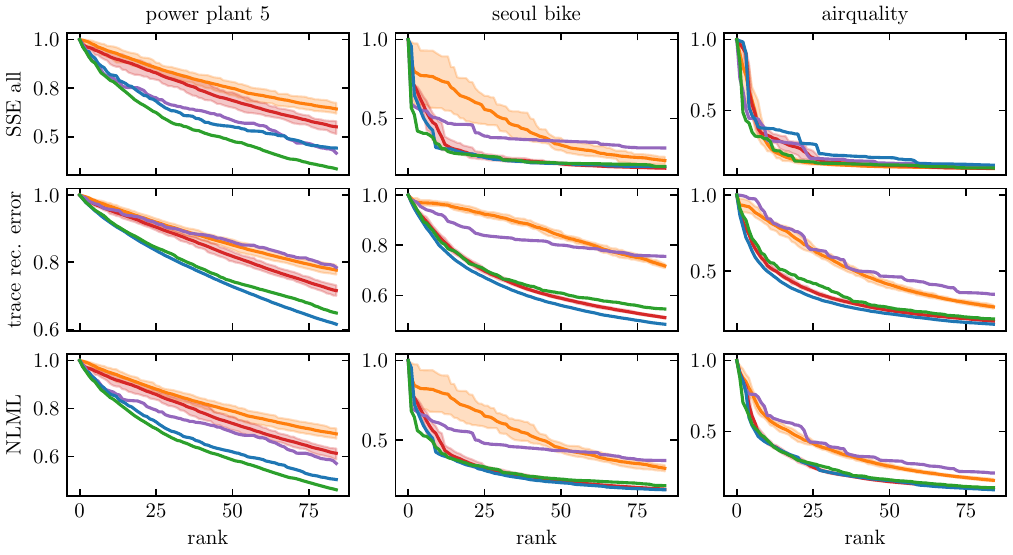}\\
    \includegraphics{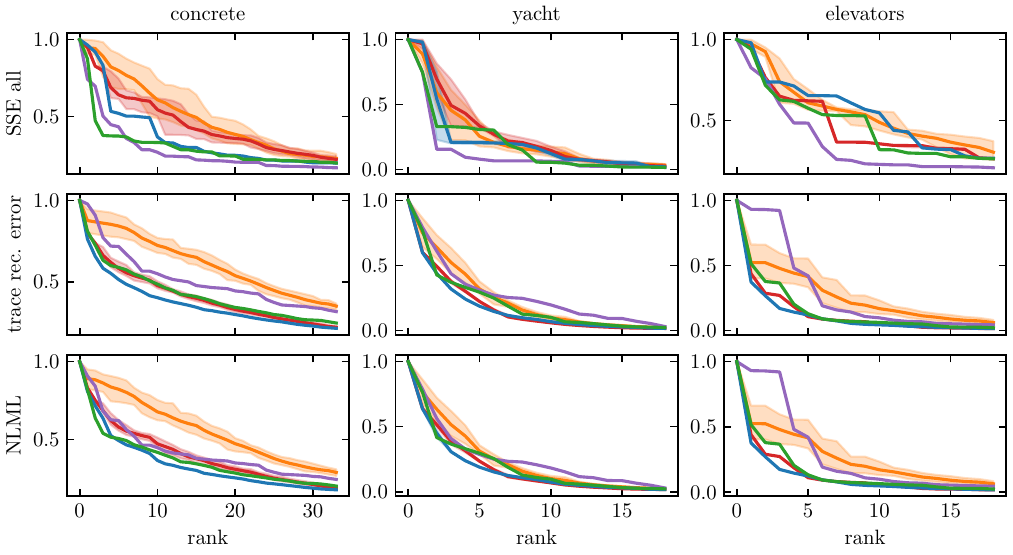}
	\caption{(continued)}
\end{figure*}

\end{document}